%% file: main.tex
\documentclass[10pt,twocolumn,letterpaper]{article}

\usepackage[pagenumbers]{iccv}
\usepackage[accsupp]{axessibility}
\usepackage[utf8]{inputenc}
\usepackage[T1]{fontenc}
\usepackage{url}
\usepackage{booktabs}
\usepackage{amsfonts}
\usepackage{nicefrac}
\usepackage{microtype}
\usepackage{xcolor}
\usepackage{colortbl}
\usepackage{graphicx}

\usepackage{amsmath}
\usepackage{amsthm}
\usepackage{pgfplots}
\usepackage{amssymb}
\pgfplotsset{compat=1.9}
\usepackage{subcaption}
\usepackage{caption}
\usepackage{tikz}
\usetikzlibrary{arrows.meta, positioning}
\usepackage{wrapfig}
\usepackage{bm}
\usepackage{algorithm}
\usepackage{algpseudocode}
\usepackage{array}
\usepackage{multirow}
\usepackage{pifont}
\usepackage{tikz}
\usepackage{bm}

\usetikzlibrary{patterns,calc,decorations.pathreplacing,shapes,arrows.meta,positioning,backgrounds,fit,matrix}

\newtheorem{theorem}{Theorem}
\newtheorem{lemma}{Lemma}

\input{preamble}
\definecolor{iccvblue}{rgb}{0.21,0.49,0.74}
\usepackage[pagebackref,breaklinks,colorlinks,allcolors=iccvblue]{hyperref}

\def\paperID{8176}
\def\confName{ICCV}
\def\confYear{2025}

\title{Backdooring Self-Supervised Contrastive Learning by Noisy Alignment}

\author{
\begin{tabular}{c}
Tuo~Chen$^{1,3}$,
Jie~Gui$^{1,2\ddagger}$,
Minjing~Dong$^{4}$,
Ju~Jia$^{1}$,
Lanting~Fang$^{5}$,
Jian~Liu$^{3\ddagger}$\\[4pt]
$^{1}$Southeast University \quad
$^{2}$Purple Mountain Laboratories \quad
$^{3}$Ant Group\\
$^{4}$City University of Hong Kong \quad
$^{5}$Beijing Institute of Technology
\end{tabular}
}

\makeatletter
\renewcommand{\@fnsymbol}[1]{\ensuremath{\ifcase#1\or *\or \dagger\or \ddagger\else\@ctrerr\fi}}
\makeatother

\begin{document}
\maketitle

\begingroup
\renewcommand\thefootnote{}\footnotetext{$^\ddagger$\,Corresponding authors: Jian Liu (\texttt{rex.lj@antgroup.com}) and Jie Gui (\texttt{guijie@seu.edu.cn}).}
\endgroup

\begin{abstract}
    Self-supervised contrastive learning (CL) effectively learns transferable representations from unlabeled data containing images or image-text pairs but suffers vulnerability to data poisoning backdoor attacks (DPCLs). An adversary can inject poisoned images into pretraining datasets, causing compromised CL encoders to exhibit targeted misbehavior in downstream tasks. Existing DPCLs, however, achieve limited efficacy due to their dependence on fragile implicit co-occurrence between backdoor and target object and inadequate suppression of discriminative features in backdoored images. We propose Noisy Alignment (NA), a DPCL method that explicitly suppresses noise components in poisoned images. Inspired by powerful training-controllable CL attacks, we identify and extract the critical objective of noisy alignment, adapting it effectively into data-poisoning scenarios. Our method implements noisy alignment by strategically manipulating contrastive learning's random cropping mechanism, formulating this process as an image layout optimization problem with theoretically derived optimal parameters. The resulting method is simple yet effective, achieving state-of-the-art performance compared to existing DPCLs, while maintaining clean-data accuracy. Furthermore, Noisy Alignment demonstrates robustness against common backdoor defenses. Codes can be found at \url{https://github.com/jsrdcht/Noisy-Alignment}.
\end{abstract}

\input{sec/intro}

\input{sec/related_work}

\input{sec/method}

\input{sec/experiment}

\section{Conclusion}
In this paper, we propose a novel data poisoning backdoor attack against contrastive learning (DPCL), where noisy backdoored images are aligned with reference images. We formulate noisy alignment as an image placement problem in 2D space and derive the optimal layout. Despite its simplicity, our method achieves state-of-the-art attack performance. Extensive experiments demonstrate that common defenses struggle to mitigate our attack effectively. Our study highlights the urgent need for more robust defenses.

\section{Acknowledgment}
This work was supported in part by the grant of the National Natural Science Foundation of China under Grant 62172090; Start-up Research Fund of Southeast University under Grant RF1028623097. We thank the Big Data Computing Center of Southeast University for providing the facility support on the numerical calculations.

{
    \small

    \bibliographystyle{ieeenat_fullname}
    \bibliography{zotero.bib}
}
\appendix

\newpage

\input{sec/supplementary.tex}

\end{document}

%% file: sec/intro.tex
\section{Introduction}
\label{sec:intro}
Self-supervised contrastive learning has revolutionized representation learning by mapping data into embedding spaces where semantic similarity correlates with proximity \cite{he2020Momentum,chen2020Simple}. Modern implementations like CLIP \cite{radford2021Learning} and DINOv2 \cite{oquab2024DINOv2} leverage web-scale datasets to achieve remarkable zero-shot generalization and have wide application potential in different downstream tasks.
However, the uncurated nature of these data introduces a significant risk of data contamination. Such datasets typically scraped from internet sources (e.g., Google, YouTube) \cite{radford2021Learning,oquab2024DINOv2}, often lack manual review before being fed into the model.
Recent studies indicate that contrastive learning is susceptible to data poisoning backdoor attacks \cite{saha2022Backdoor,carlini2021Poisoning,zhang2024Data,bansal2023CleanCLIP,li2023Embarrassingly}.
In extreme cases, it is feasible to manipulate a contrastive learning model to misclassify a backdoored test input by corrupting as little as one millionth of the pre-training dataset \cite{carlini2021Poisoning}.

DPCL exploits the co-occurrence from random augmentations of backdoor triggers and target object patterns in images \cite{carlini2021Poisoning,zhang2024Data}. Given an image, CL randomly generates augmented views and enforces similarity (dissimilarity) between features of positive (negative) views. By poisoning pre-training data with malicious images containing dog patterns and backdoor triggers, victim CL models learn to associate triggers with dogs (the attack target). Consequently, downstream classifiers inherit this bias and misclassify triggered images as "dog". Existing DPCL methods \cite{saha2022Backdoor,carlini2021Poisoning,zhang2024Data,bansal2023CleanCLIP,li2023Embarrassingly} universally leverage this principle. For instance, Saha et al. \cite{saha2022Backdoor} physically superimpose triggers onto targets, while Zhang et al. \cite{zhang2024Data} optimize co-occurrence probabilities. This paper focuses on image-modal CL, with Section \ref{sec:extension_to_image_text_CL} extending our approach to image-text CL.

\begin{figure}
  \centering
  \includegraphics[width=\linewidth]{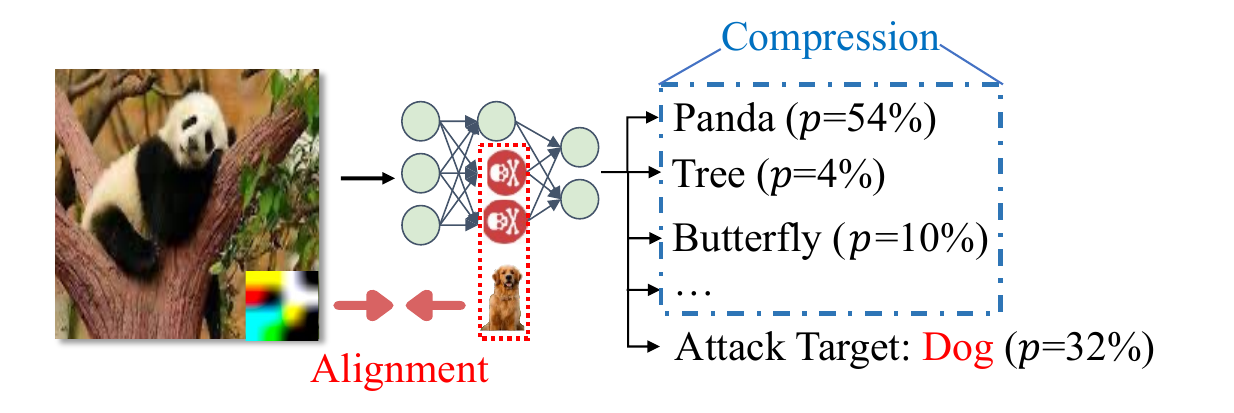}
  \vspace{-0.3in}
  \caption{Illustration of our intuition.}
  \vspace{-0.2in}
  \label{fig:figure_striking}
\end{figure}

Current DPCLs exhibit limited attack effectiveness. To bridge this gap, we draw inspiration from a theoretical upper bound backdoor attack to CL (called \textit{oracle attack}) that controls model training \cite{jia2022BadEncoder,tao2023Distribution,wang2024TransTroj,wang2023GhostEncoder}. Oracle attack essentially maximizes the feature similarity between reference images (collected target-class images guiding the attack) and noisy backdoored images. By decomposing the oracle attack objective, i.e., \textit{noisy alignment}, into representation-space reference alignment components that capture the co-occurrence of backdoor and target object patterns and noise compression components that capture the degradation of original noisy patterns, we demonstrate that the noise compression term inherently compresses the subspace orthogonal to reference features. As illustrated in Figure \ref{fig:figure_striking}, backdoored panda images may fail due to domination by non-trigger features. Enhancing attack performance requires suppressing the neural network's extraction of undesirable elements (e.g., pandas or trees) beyond the backdoor trigger. Formal analysis appears in Section \ref{sec:problem_formulation}. Existing DPCLs only consider the alignment component, lacking the compression component, which we hypothesize leads to their limited attack efficacy.

Oracle attacks require control over the training process, which becomes infeasible in data poisoning scenarios. \textbf{Our objective is to approximate oracle attack effectiveness under practical data poisoning constraints.} Noisy alignment can be simulated by treating augmented views of both image types as positive pairs. If one augmented view contains (a part of) a noisy backdoored image and the other contains (a part of) a reference image, the CL model would produce similar features for both views.
To this end, we propose a novel DPCL method, termed NA (\textbf{\textit{N}}oisy \textit{\textbf{A}}lignment), explicitly achieving the reference alignment and noise compression objectives of oracle attacks by manipulating the random cropping augmentation. Our method introduces two key innovations to address existing DPCL limitations. (1) We explicitly formulate noise compression as a part of the attack objective. This is achieved by collecting a small set of images and converting them into backdoored noisy images. This compels CL encoders to suppress discriminative features orthogonal to the attack target, thereby amplifying trigger effectiveness. (2) We devise an offline, optimal poison crafting strategy to achieve noisy alignment under data poisoning scenarios. Our method inverts the random cropping in CL, ensuring poison images' random crops capture either noisy or reference images. To maximize the probability of satisfying these conditions simultaneously, we model poison crafting as a two-dimensional layout optimization problem between reference and backdoored noisy image regions and theoretically derive optimal crafting parameters.

We compare our method with existing DPCLs on different datasets and CL models.
Our experiments show that Noisy Alignment achieves state-of-the-art performance, with ASR improvements ranging from 1.2\% to 45.9\% on ImageNet-100, while keeping the utility on the clean data.
Our Noisy Alignment can be easily adapted to image-text contrastive learning.
Additionally, we evaluated potential defenses, including supervised methods, those tailored for self-supervised learning, and our own adaptive defense. We demonstrate that both supervised and self-supervised backdoor detection methods struggle to detect our attack. Our adaptive defense nullifies the backdoor by disrupting the malicious co-occurrence, further validating the core intuition of our approach.

Our contributions are outlined as follows:
\begin{itemize}
    \item We propose a new DPCL objective called Noisy Alignment, which explicitly
    approximates powerful oracle attacks in data poisoning scenarios.
    \item We develop a poisons crafting strategy to get the optimal poisons layout to achieve Noisy Alignment.
    \item We validate the effectiveness of Noisy Alignment through extensive experiments.
\end{itemize}

%% file: sec/related_work.tex
\section{Related Work}

\subsection{Data Poisoning-based Backdoor Attacks to Self-Supervised Contrastive Learning}

Generally, an adversary augments the original dataset with poisoning samples that contain a trigger in order to induce the model trained on this dataset to behave incorrectly.
SSLBKD \cite{saha2022Backdoor} naively embeds triggers into the target class samples.
CTRL \cite{li2023Embarrassingly} proposed using frequency-domain backdoor to enhance backdoor stealthiness. PoisonedEncoder~\cite{liu2022PoisonedEncoder} explored backdoor attacks to CL under a targeted poisoning setup.
CorruptEncoder~\cite{zhang2024Data} carefully placing the trigger to maximize the probability that the interested object co-occurs with the trigger.
BLTO \cite{sun2023Backdoor} crafts the dynamic trigger by training a generative covolutional neural network.
Li et al.~\cite{li2024Difficulty} show that DPCL entangles backdoor features with those of the target class, making defense more difficult.
Another line of research \cite{tao2023Distribution,xue2023ESTAS,jia2022BadEncoder,wang2024TransTroj,wang2023GhostEncoder,zhang2024Imperceptible} focuses on backdooring pre-trained SSL encoders.

\subsection{Noise in Self-Supervised Learning}

Noise undermines self-supervised learning by degrading representation quality \citep{amrani2021Noise}. However, tackling this noise may improve outcomes.
Denoising itself can be supervision, \cite{lehtinen2018Noise2Noise,batson2019noise2self} train denoising models with paired noisy observations.
InfoMin \cite{tian2020What} suggests that models can be encouraged to compress
excess noise in data. The noisy views and mismatched pairs that commonly arise in large-scale or multimodal SSL can be explicitly modeled. \cite{chuang2020Debiased} corrects the bias from false negatives in InfoNCE using a PU-learning view. For misaligned video-text pairs, MIL-NCE \cite{miech2020end} uses multiple-instance matching to tolerate temporal misalignment, while Robust Audio-Visual Instance Discrimination \cite{morgado2021robust} reweights false positives/negatives across modalities.

%% file: sec/method.tex
\section{Preliminaries}
\label{sec:preliminaries}

In this section, we introduce our threat model and notations.
Following previous work \cite{saha2022Backdoor,zhang2024Data,li2023Embarrassingly}, we take the image classification as the downstream task for clarity.

\noindent\textbf{Data poisoning in Self-supervised contrastive learning.} Suppose the original pre-training dataset is $\mathcal{D}_{\text{pr}} \subset \mathcal{X}$ where $\mathcal{X}$ is the image space.
A victim trains an encoder $f_{\theta}: \mathcal{X} \rightarrow \mathbb{R}^d$
on $\mathcal{D}_{\text{pr}}$ with contrastive loss $\mathcal{L}_{\text{cl}}: \mathbb{R}^d \times \mathbb{R}^d \rightarrow  \mathbb{R}$ to learn representations.
After that, the downstream users train a downstream classifier based on the representation from the infected encoder to perform any given downstream task. Let $\theta_f$ be the parameters of the encoder.
For a specific interested downstream task, the adversary injects a corresponding small set of poisons $\mathcal{D}_{\text{p}} \subset \mathcal{X}$ into the pre-training data to mislead the downstream classifier built on the pre-trained infected encoder $f_{\hat{\theta}_f}$ to incorrectly classify poisoned examples as the pre-defined target class $t$.
In this paper, hat notation $\hat{\cdot}$ denotes the infected version of the original variable.

\noindent\textbf{Adversary's knowledge and capability}: Similar to previous work \cite{zhang2024Data,saha2022Backdoor,li2023Embarrassingly}, the adversary can collect a small reference set $\mathcal{D}_{\text{ref}} \subset \mathcal{X}$ corresponding to the interested class $t$ to guide the poisoning process and inject a small set of poisons $\mathcal{D}_{\text{p}}$ into the training data, e.g., $\frac{|\mathcal{D}_{\text{p}}|}{|\mathcal{D}_{\text{pr}}|} \leq 0.5\%$.
Apart from this, we assume that the adversary has access to a small subset $\mathcal{D}_{\text{shadow}}$ of the reference distribution.
The adversary lacks insight into (i) the model details (e.g., network architectures or CL methods) and (ii) detailed training settings (e.g., optimizers or learning rate schedulers).

\begin{figure}[t]
    \centering
    \begin{subfigure}[b]{0.49\columnwidth}
        \centering
        \includegraphics[width=\textwidth]{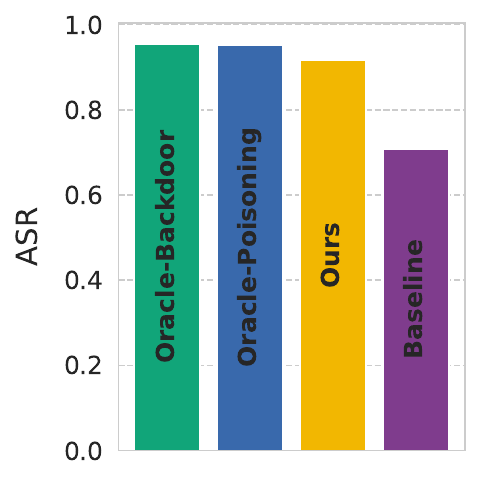}
        \caption{ASR comparison}
        \label{fig:asr_comparison_between_variants}
    \end{subfigure}
    \hfill
    \begin{subfigure}[b]{0.49\columnwidth}
        \centering
        \includegraphics[width=\textwidth]{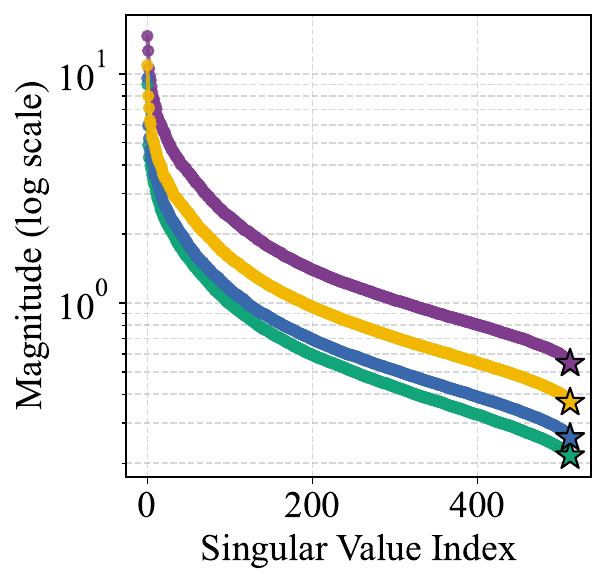}
        \caption{Representation rank}
        \label{fig:representation_rank}
    \end{subfigure}
    \caption{Comparison of different DPCL variants.
    (a) ASR of different DPCL variants. (b) Singular value distribution of representation matrix. Smaller singular values indicate reduced rank and collapse in the space.}
    \vspace{-0.1in}
\end{figure}

\section{Improving DPCLs by Compressing Noise}
\label{sec:problem_formulation}

As shown in Table \ref{tab:targeted_performance}, existing DPCLs \cite{saha2022Backdoor,li2023Embarrassingly,zhang2024Data} lag far behind training-controllable self-supervised contrastive learning backdoor attacks \cite{jia2022BadEncoder,tao2023Distribution} in terms of attack performance.
In this section, we analyze the reasons behind this phenomenon and explore ways to improve DPCLs to bridge the gap. Since training-controllable methods represent the upper bound of DPCL performance, we refer to them as oracle attacks.
The oracle attack can be formulated as the malicious objective below:
\begin{equation}
    \min_{\theta_f} \mathbb{E}_{\substack{
        \mathbf{x} \sim \mathcal{D}_{\text{pr}} \
        }} [\mathcal{L}_{\text{cl}}] + \underbrace{ \mathbb{E}_{\substack{\mathbf{x}_{\text{s}} \sim \mathcal{D}_{\text{shadow}} \\ \mathbf{x}_{\text{r}} \sim \mathcal{D}_{\text{ref}}}}  \left[1 - \cos\left(f(\mathbf{x}_{\text{s}} \oplus \mathbf{p}), f(\mathbf{x}_{\text{r}})\right)\right]}_{\text{noisy alignment loss}\ \mathcal{L}_{\text{align}}}
    \label{eq:oracle_objective}
\end{equation}
\noindent where $\mathcal{L}_{\text{align}}$ enforces that board infected shadow examples $\hat{\mathbf{x}_{\text{s}}} \!=\! \mathbf{x}_{\text{s}} \oplus \mathbf{p}$ align with reference examples via cosine similarity.
$\oplus$ is the trigger embedding operation which is typically used to embed a backdoor trigger $\mathbf{p} \in \mathcal{X}$ into any victim image $\mathbf{x}$ to craft an infected version $\hat{\mathbf{x}}$.
Specifically, for each poisoned shadow image during training, we randomly select a reference image $\mathbf{x}_{\text{r}} \!\sim\! \mathcal{D}_{\text{ref}}$ and minimize their feature distance in hyperspherical space.
Objective \eqref{eq:oracle_objective} is from \cite{jia2022BadEncoder} and simplifies the loss terms that are unrelated to the attack.

\begin{figure}[t]
    \centering
    \centering
    \includegraphics[width=\columnwidth]{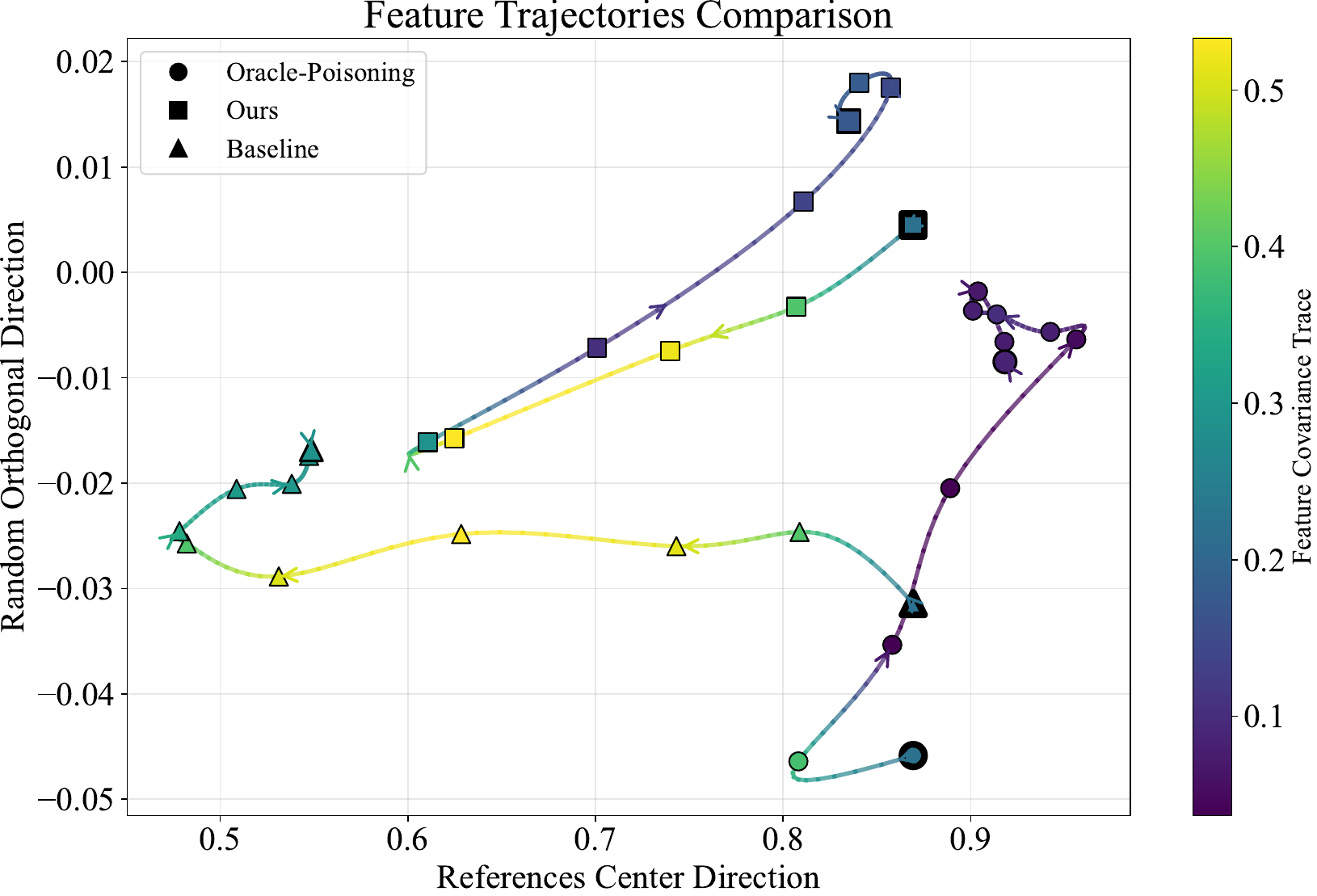}
    \caption{2D projection visualization of the training trajectories.
    Bold markers indicate the start point, and arrows indicate the training direction. Darker colors represent smaller traces of the feature covariance matrix, indicating stronger collapse in the $\mathbf{v}_\perp$ space defined in Equation \eqref{eq:decomposition}.}
    \label{fig:training_trajectories}
    \vspace{-0.1in}
\end{figure}

Intuitively, the noisy alignment term performs the attack by projecting malicious samples into the feature neighborhood of the target class. However, we demonstrate that, in addition to enforcing reference alignment, the noisy alignment loss implicitly accomplishes the task of noise compression.
Specifically, by decomposing the features of any $\mathbf{x}_{\text{s}} \oplus \mathbf{p}$ in the hyperspherical space, we reveal implicit geometric constraints in $\mathcal{L}_{\text{align}}$. Let $f(\mathbf{x}_{\text{r}}) = \mathbf{u}$ denote the unit-norm reference feature (L2-normalized as per contrastive learning convention), and $f(\mathbf{x}_{\text{s}} \oplus \mathbf{p}) = \mathbf{v}$ be the poisoned feature. We decompose $\mathbf{v}$ into two orthogonal components:
\begin{equation}
    \label{eq:decomposition}
    \mathbf{v} = \underbrace{(\mathbf{v}^\top \mathbf{u})\mathbf{u}}_{\text{Alignment component}} + \underbrace{\mathbf{v}_\perp}_{\text{Compression component}},
\end{equation}
where $\mathbf{v}_\perp = \mathbf{v} - (\mathbf{v}^\top \mathbf{u})\mathbf{u}$ represents the residual component orthogonal to $\mathbf{u}$. The cosine similarity term in $\mathcal{L}_{\text{align}}$ then be formulated as:
\begin{equation}
    \cos(\mathbf{v}, \mathbf{u}) = \frac{\mathbf{v}^\top \mathbf{u}}{\|\mathbf{v}\|} = \frac{\alpha}{\sqrt{\alpha^2 + \|\mathbf{v}_\perp\|^2}}, \nonumber
\end{equation}
\noindent where $\alpha = \mathbf{v}^\top \mathbf{u}$. Substituting this into $\mathcal{L}_{\text{align}}$, we get:
\begin{equation}
    \mathcal{L}_{\text{align}} = \mathbb{E}\left[1 - \frac{\alpha}{\sqrt{\alpha^2 + \|\mathbf{v}_\perp\|^2}}\right]. \nonumber
\end{equation}
This formulation reveals two implicit objectives:
\begin{itemize}
    \item \textbf{Alignment Term}: Maximizing $\alpha$ to increase the projection of poisoned features onto the reference direction $\mathbf{u}$;
    \item \textbf{Compression Term}: Minimizing $\|\mathbf{v}_\perp\|^2$ to suppress features orthogonal to $\mathbf{u}$, effectively compressing the variance of poisoned samples' features.
\end{itemize}
The gradient dynamics confirm this decomposition. The gradient of $\mathcal{L}_{\text{align}}$ with respect to $\alpha$ and $\|\mathbf{v}_\perp\|^2$ becomes:
\begin{align}
    \frac{\partial \mathcal{L}_{\text{align}}}{\partial \alpha} &\propto -\frac{\|\mathbf{v}_\perp\|^2}{(\alpha^2 + \|\mathbf{v}_\perp\|^2)^{3/2}}, \nonumber \\
    \frac{\partial \mathcal{L}_{\text{align}}}{\partial \|\mathbf{v}_\perp\|^2} &\propto \frac{\alpha}{2(\alpha^2 + \|\mathbf{v}_\perp\|^2)^{3/2}}. \nonumber
\end{align}
\noindent These gradients simultaneously push $\alpha \to +\infty$ (perfect alignment) and $\mathbf{v}_\perp \to \mathbf{0}$ (dimensional collapse). Consequently, the poisoned features cluster tightly around $\mathbf{u}$, discarding their original discriminative features from $\mathbf{x}_{\text{s}}$.
This dual mechanism explains why simple alignment losses can achieve effective backdoor implantation. The compression effect prevents poisoned features from dispersing across the embedding space. Consider $\mathbb{E}[f(\mathbf{x}_{\text{s}} \oplus \mathbf{p})] = \mathbb{E}[f(\mathbf{p})] + \mathbb{E}[f(\mathbf{x}_{\text{s}}) + \text{Risidual Terms}]$,
noise compression forces the shadow features and Risidual Terms vectors to be collapsed into null space of noisy alignment loss since they are noisy and hard to align with the reference features.

\begin{figure}[t]
    \centering
    \includegraphics[width=\columnwidth]{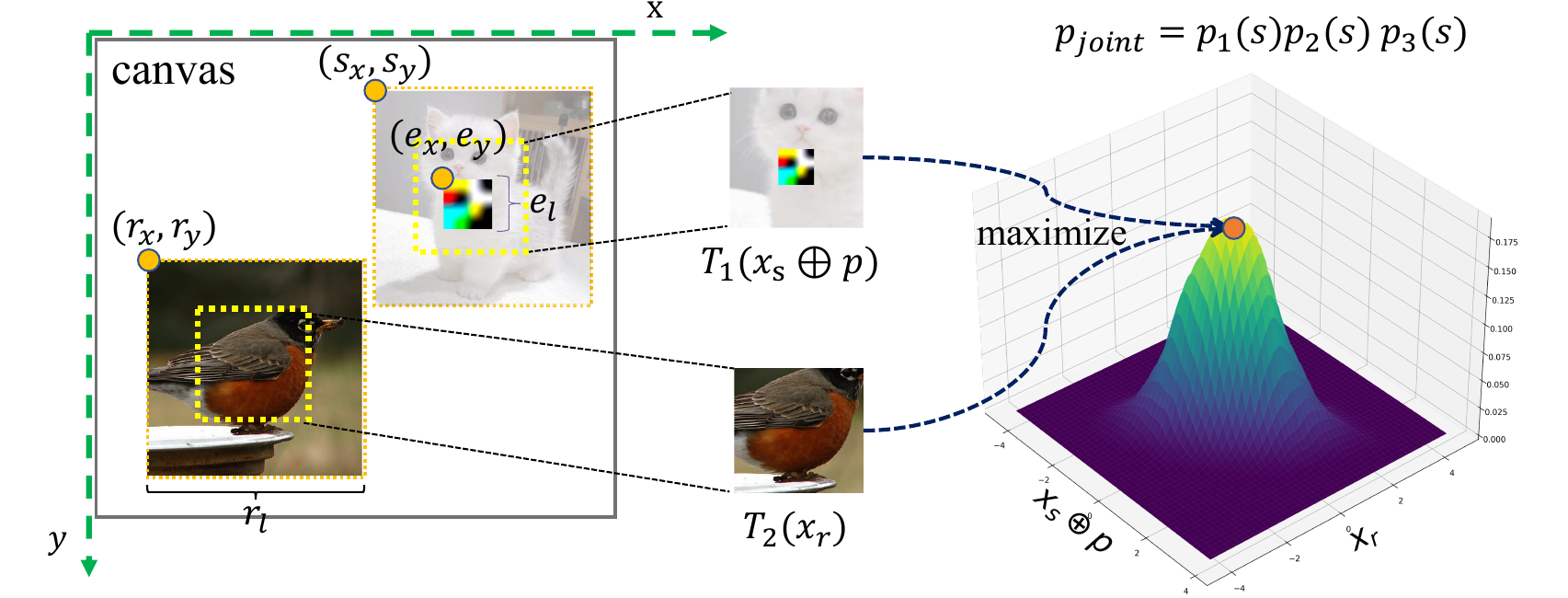}
    \caption{Maximizing likelihood of joint probability.}
    \label{fig:maximize_likelihood}
    \vspace{-0.1in}
\end{figure}

Building on the insight above, we design a data poisoning variant that integrates noisy alignment constraints into contrastive learning.
For each reference sample $\mathbf{x}_{\text{r}} \in \mathcal{D}_{\text{ref}}$, we generate two augmented views:
1) reference view $T_{1}(\mathbf{x})$ 2) shadow view $T_{2}(\mathbf{x}_{\text{s}} \oplus \mathbf{p})$ where $\mathbf{x}_{\text{s}} \sim \mathcal{D}_{\text{shadow}}$. $T_1, T_2 \overset{i.i.d.}{\sim} \mathcal{T}$ where $\mathcal{T}$ is the CL augmentation distribution.
For each batch containing clean pairs $(\mathbf{x}, \mathbf{x})\ \text{where}\ x \!\sim\! \mathcal{D}_{\text{pr}}$ and malicious pairs $(\mathbf{x}_{\text{s}}, \mathbf{x}_{\text{r}}) \!\sim\! \mathcal{D}_{\text{shadow}} \!\times\! \mathcal{D}_{\text{ref}}$, we define the oracle poisoning variant as
\begin{align}
    &\mathcal{L}_{\text{oracle-poisoning}} = \min_{\theta_f} \mathbb{E}_{\substack{
    \mathbf{x} \sim \mathcal{D}_{\text{pr}} \
    }}  [\mathcal{L}_{\text{cl}}] + \nonumber \\
    &\mathbb{E}_{\substack{
    \mathbf{x}_{\text{s}} \sim \mathcal{D}_{\text{shadow}} \\
    \mathbf{x}_{\text{r}} \sim \mathcal{D}_{\text{ref}} \
    }} \Big[ \mathcal{L}_{\text{cl}}\big(f(T_1(\mathbf{x}_{\text{s}} \oplus \mathbf{p})), f(T_2(\mathbf{x}_{\text{r}}))\big) \Big].
\end{align}
\noindent The variant enforces alignment between shadow-reference pairs while maintaining the form of contrastive learning.

\textbf{Discussion.} Figure \ref{fig:asr_comparison_between_variants} demonstrates that oracle poisoning variant matches the ASR of the oracle attack.
The baseline is from \cite{saha2022Backdoor}. Geometric analyses of the training dynamics of infected samples in Figures \ref{fig:representation_rank}-\ref{fig:training_trajectories} confirm the same observation: poisoned representations collapse orthogonally to the reference direction $\mathbf{u}$ (formalized in Eq. \eqref{eq:decomposition}).
Despite its effectiveness, oracle poisoning requires real-time access to training batches for generating malicious pairs $(\mathbf{x}_{\text{s}} \oplus \mathbf{p}, \mathbf{x}_{\text{r}})$. This violates our data poisoning threat model. We next eliminate this dependency by reformulating the noise-compression effects into static constraints pre-computable on $\mathcal{D}_{\text{p}}$.

\begin{figure}[t]
    \centering
    \begin{tikzpicture}

        \definecolor{refcolor}{RGB}{70,130,180}
        \definecolor{shadowcolor}{RGB}{210,210,210}
        \definecolor{triggercolor}{RGB}{220,50,50}
        \definecolor{dotcolor}{RGB}{255,165,0}
        \definecolor{axiscolor}{RGB}{40,40,40}

        \tikzset{
          box/.style={rectangle, draw, minimum width=1.8cm, minimum height=1.8cm},
          smallbox/.style={rectangle, draw, minimum width=0.4cm, minimum height=0.4cm},
          label/.style={font=\small},
          arrow/.style={-{Stealth[length=2mm]},thick},
          dashedarrow/.style={-{Stealth[length=2mm]},thick,dashed},
          dot/.style={circle, fill=dotcolor, inner sep=1.5pt},
          axis/.style={-{Stealth[length=3mm]}, thick, axiscolor}
        }

        \begin{scope}

          \draw[axis] (0,2) -- (5,2) node[right] {$x$};
          \draw[axis] (0,2) -- (0,-0.5) node[below] {$y$};

          \filldraw[white, draw=axiscolor] (0,2) circle (0.07);
          \node[label, anchor=north west] at (0,2) {$O$};

          \foreach \x in {1,2,3,4}
            \draw (\x,2) -- (\x,1.9);
          \foreach \y in {1,0,-0.3}
            \draw (0,\y) -- (0.1,\y);

          \draw[thick] (0,0) rectangle (4,2);
          \node[label, anchor=south east] at (4,0) {$c_w^*=2r_l$};
          \node[label, anchor=south west] at (4,0.7) {$c_h^*=r_l$};

          \filldraw[refcolor!40, draw=black] (0,0) rectangle (2,2);
          \node[label] at (1,1) {$\mathbf{x}_{\text{r}}$};

          \node[dot] (r_point) at (0,2) {};
          \node[label] (r_label) at (1.2,2.5) {$(r_x^*,r_y^*)=(0,0)$};
          \draw[arrow] (r_label) -- (r_point);

          \filldraw[shadowcolor!80, draw=black] (2,0) rectangle (4,2);
          \node[label] at (3,1.5) {$\mathbf{x}_{\text{s}} \oplus \mathbf{p}$};

          \node[dot] (s_point) at (2,2) {};
          \node[label] (s_label) at (3.9,2.3) {$(s_x^*,s_y^*)=(r_l,0)$};
          \draw[arrow] (s_label) -- (s_point);

          \filldraw[triggercolor!80, draw=black] (2.75,0.75) rectangle (3.25,1.25);
          \node[label, white] at (3,1) {$\mathbf{p}$};

          \node[dot] (e_point) at (2.75,1.25) {};
          \node[label] (e_label) at (2,0.5) {$(e_x^*,e_y^*)$};
          \draw[arrow] (e_label) -- (e_point);

          \draw[<->] (0,-0.1) -- (2,-0.1);
          \node[label] at (1,-0.6) {$r_l$};
          \draw[<->] (2,-0.1) -- (4,-0.1);
          \node[label] at (3,-0.6) {$r_l$};
        \end{scope}

    \end{tikzpicture}
    \caption{Optimal parameters for left-right layout according to Theorem \ref{thm:locations} and \ref{thm:canvas_size}.}
    \label{fig:optimal_parameters}
    \vspace{-0.1in}
\end{figure}
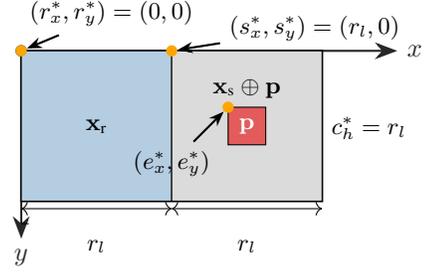

\section{Offline Noise Compression for DPCL}
We reformulate the noisy alignment as a static data perturbation by interpreting the backdoor implantation as a adaptive inverse of CL augmentation.
Let $\mathcal{T}_\oplus: (\mathbf{x}_{\text{s}} \oplus \mathbf{p}, \mathbf{x}_{\text{r}}) \mapsto \widehat{\mathbf{x}_{\text{s}},\mathbf{x}_{\text{r}}}$ where $\widehat{\mathbf{x}_{\text{s}},\mathbf{x}_{\text{r}}} \in \mathcal{X}$ is the composite image denotes our poisoning function that combines trigger-embedded shadow images and reference images. To simulate oracle poisoning's dynamics without training access, we pre-optimize $\mathcal{T}_\oplus$ to maximize the likelihood of any malicious pair $(\mathbf{x}_{\text{s}} \oplus \mathbf{\text{p}}, \mathbf{x}_{\text{r}})$ being treated as positive pairs in contrastive learning.
Specifically, we define the objective as
\begin{align}
    \label{eq:maximize_likelihood}
    &\max_{\mathcal{T}_\oplus} \mathbb{E}_{\substack{
    T_1,T_2 \overset{\text{i.i.d.}}{\sim} \mathcal{T} \\
    \mathbf{x}_{\text{s}}, \mathbf{x}_{\text{r}} \sim \mathcal{D}_{\text{shadow}} \times \mathcal{D}_{\text{ref}}
    }}
    \left[ \Pr\left(
    T_1(\widehat{\mathbf{x}_{\text{s}},\mathbf{x}_{\text{r}}}) \in \mathcal{A}(\mathbf{x}_{\text{s}} \oplus \mathbf{p}) \right. \right. \nonumber \\
    &\left. \left. \quad \wedge  \quad
    T_2(\widehat{\mathbf{x}_{\text{s}},\mathbf{x}_{\text{r}}}) \in \mathcal{A}(\mathbf{x}_{\text{r}})
    \right) \right]
\end{align}
\noindent
where $\mathcal{A}(\cdot)$ denotes the augmentation neighborhood of an input.
This objective ensures that random augmentations of the composite sample preserve both the trigger pattern from $\mathbf{x}_{\text{s}} \oplus \mathbf{p}$ and discriminative features from $\mathbf{x}_{\text{r}}$.
However, the expectation of joint probability is intractable due to the inability to access the victim's data augmentation process. Following observations in \cite{chen2020Simple,zhang2024Data}, random cropping dominates CL poisoning. We thus simplify the joint probability by decoupling it into independent events as
\begin{equation}
\label{eq:maximize_joint_likelihood}
        \Pr\left(
        \underbrace{(\mathbf{p} \subseteq \mathcal{V}_1 \subseteq \mathbf{x}_{\text{s}} \oplus \mathbf{p})}_{\text{trigger retention}} \! \wedge \!
        \underbrace{(\mathcal{V}_2 \subseteq \mathbf{x}_{\text{r}})}_{\text{reference matching}} \! \wedge \!
        \underbrace{(\mathcal{V}_1 \cap \mathcal{V}_2 = \emptyset)}_{\text{view disjoint}},
        \right)
\end{equation}
\noindent where $\mathcal{V}_1 = T_1(\widehat{\mathbf{x}_{\text{s}},\mathbf{x}_{\text{r}}}), \mathcal{V}_2=T_2(\widehat{\mathbf{x}_{\text{s}},\mathbf{x}_{\text{r}}})$.
This reduces the Equation \eqref{eq:maximize_likelihood} to
a practical 2D layout optimization problem under random cropping distributions.
We demonstrate our intuition in Figure \ref{fig:maximize_likelihood}.
We enforce spatial disjointness to prevent information leakage through that would enable models to bypass contrastive optimization via shortcut \citep{tian2020What}.
The adversary needs to maximize the likelihood of trigger retention and reference matching by carefully designing the layout of the composite image $\mathcal{T}_\oplus (\mathbf{x}_{\text{s}} \oplus \mathbf{p}, \mathbf{x}_{\text{r}})$.

\begin{algorithm}[t]
    \caption{Crafting Poisoned Dataset}
    \label{alg:method}
    \begin{algorithmic}[1]
        \Require Backdoor trigger $\mathbf{p}$, reference set $\mathcal{D}_{\text{ref}}$, shadow set $\mathcal{D}_{\text{shadow}}$
        \Ensure Poisoned dataset $\mathcal{D}_\text{p}$
        \State Initialize poisoned dataset $\mathcal{D}_\text{p} \leftarrow \emptyset$
        \While{not converged}
            \State Sample reference image $\mathbf{x}_\text{r} \sim \mathcal{D}_{\text{ref}}$
            \State Sample shadow image $\mathbf{x}_\text{s} \sim \mathcal{D}_{\text{shadow}}$
            \State Embed trigger into shadow image: $\mathbf{x}_\text{s} \oplus \mathbf{p}$
            \State Sample layout direction from \{left-right, right-left, up-down, down-up\}
            \State Determine optimal parameters based on Theorems 1 \& 2:
                \State \quad Set reference position $(r_x^*, r_y^*)$, shadow position $(s_x^*, s_y^*)$, trigger position $(e_x^*, e_y^*)$ at center of shadow image, canvas size $(c_w^*, c_h^*)$
                \State \quad Create composite image $\widehat{\mathbf{x}_{\text{s}},\mathbf{x}_{\text{r}}} = \mathcal{T}_\oplus(\mathbf{x}_\text{s} \oplus \mathbf{p}, \mathbf{x}_\text{r})$ based on layout
            \State Update poisoned dataset: $\mathcal{D}_\text{p} \leftarrow \mathcal{D}_\text{p} \cup \{\widehat{\mathbf{x}_{\text{s}},\mathbf{x}_{\text{r}}}\}$
        \EndWhile
        \State \Return $\mathcal{D}_\text{p}$
    \end{algorithmic}
\end{algorithm}

Formally, we denote by $\mathbf{x}_{\text{r}}$ the reference image, $\mathbf{x}_{\text{s}}$ the shadow image, $\mathbf{p}$ the trigger and $T_1, T_2$ are random cropping operations independently and identically distributed in $\mathcal{T}$.
We define the layout optimization problem as inserting the trigger $\mathbf{p}$ into the shadow image $\mathbf{x}_{\text{s}}$ and inserting the $\mathbf{x}_{\text{r}}, \mathbf{x}_{\text{s}} \oplus \mathbf{p}$ into a 2D canvas to maximize the likelihood defined in Equation \eqref{eq:maximize_joint_likelihood}.
The size of the reference image $(r_l, r_l)$ and the size of the trigegr $e_l$ are frozen, and 1) the location of the reference image $(r_x, r_y)$ 2) the location of the trigger $(e_x, e_y)$ 3) the location of the shadow image $(s_x, s_y)$ 4) the canvas size $(c_w, c_h)$ are all variables to be optimized.
To simplify the problem, we assume that the reference image, shadow image, trigger are all square and the shadow image share the same size with the reference image.

Assuming the cropped regions are squares and they have the same size $s$ (the conclusion holds if the cropped regions have different sizes). We denote by $p_1 (s)$ the probability of a randomly cropped view containing the trigger and within the infected shadow image, and $p_2 (s)$ the probability of a randomly cropped view is within the reference image.
The reference image and the infected shadow image are expected to be disjoint.
Following the formulation in \cite{zhang2024Data}, we cast the objective \eqref{eq:maximize_joint_likelihood} as the following maximization problem:
\begin{equation}
\label{eq:layout_optimization}
    p_{\text{joint}} = \frac{1}{S-e_l} \int\limits_{s \in (e_l,S]} p_1(s) p_2(s) p_3(s) \, ds.
\end{equation}
\noindent where $p_1 (s) = \Pr \{ \left(\mathbf{p} \subseteq  \mathcal{V}_1 \right)
\wedge \left(  \mathcal{V}_1  \subseteq \left( \mathbf{x}_{\text{s}} \oplus \mathbf{p} \right)  \right)  \}$, $p_2 (s) = \Pr\{   \mathcal{V}_2 \subseteq \mathbf{x}_{\text{r}} \}$ and $p_3 (s) = \Pr\{ \mathcal{V}_1 \cap \mathcal{V}_2 = \emptyset \}$. The optimizable parameters for Objective \eqref{eq:layout_optimization} include $r_x, r_y, s_x, s_y, e_x, e_y, c_w, c_h$.
The region size $s$ is uniformly distributed in the range $(e_l, S]$.

Depending on the relative positions of the reference image and the infected shadow image, there are four possible layout categories: 1) \textit{left-right}, 2) \textit{right-left}, 3) \textit{up-down}, and 4) \textit{down-up}. For example, a left-right layout indicates that the reference image is positioned to the left of the infected shadow image, meaning a vertical line can separate the two images. Different layouts can be achieved through rotational symmetry (or flipping), thus we primarily focus on the left-right layout. When generating a poisoned image, we randomly choose one of these four layouts.

\begin{theorem}[Locations of Reference Image, Trigger and Shadow Image]
\label{thm:locations}
Suppose the left-right layout is used. For any $c_h \geq r_l, c_w \geq 2r_l$, the following locations maximize the likelihood in Equation \eqref{eq:layout_optimization}.
$(r_x^*, r_y^*) = (0, 0)$ is the optimal location of the reference image.
$(s_x^*, s_y^*) = (\frac{c_w}{2}, 0)$ with $s_x \geq 2r_l$ is the optimal location of the infected shadow image.
The optimal location of the trigger is the center of the infected shadow image, i.e., $(e_x^*, e_y^*) = \left(s_x^* + \frac{r_l - e_l}{2}, s_y^* + \frac{r_l - e_l}{2}\right)$.
\end{theorem}
\begin{proof}
    See Appendix A.
\end{proof}

\begin{theorem}[Canvas Size]
\label{thm:canvas_size}
Suppose the left-right layout and the optimal locations in Theorem \ref{thm:locations} are used. For any width $c_w \geq 2r_l$, the optimal canvas height is $c_h^* = r_l$. For height $c_h = r_l$, the optimal canvas width is $c_w^* = 2r_l$.
\end{theorem}
\begin{proof}
    See Appendix A.
\end{proof}

Theorem \ref{thm:locations} and \ref{thm:canvas_size} analytically derive the optimal parameters of the left-right layout which is shown in Figure \ref{fig:optimal_parameters}.
For other layouts, the optimal parameters can be derived similarly.
Algorithm \ref{alg:method} summarizes the poison crafting process.

%% file: sec/experiment.tex
\begin{table*}[t]
    \centering
    \scriptsize
    \caption{Effectiveness of attacks on different datasets. Bold indicates the highest ASR value, and underline indicates the second highest. CTRL-NG refers to CTRL without Gaussian blur augmentation.
    BLTO-N normalizes the BLTO ASR by the ASR of the uninfected model.}
    \begin{tabular}{cccccccccccccc}
        \toprule
        \textbf{Dataset} & \textbf{Attack} & \multicolumn{3}{c}{\textbf{MoCo v2}} & \multicolumn{3}{c}{\textbf{BYOL}} & \multicolumn{3}{c}{\textbf{SimSiam}} & \multicolumn{3}{c}{\textbf{SimCLR}} \\
        \cmidrule(lr){3-5} \cmidrule(lr){6-8} \cmidrule(lr){9-11} \cmidrule(lr){12-14}
        & & \textbf{CA} & \textbf{BA} & \textbf{ASR} & \textbf{CA} & \textbf{BA} & \textbf{ASR} & \textbf{CA} & \textbf{BA} & \textbf{ASR} & \textbf{CA} & \textbf{BA} & \textbf{ASR} \\
        \midrule
        \multirow{8}{*}{ImageNet-100}
        & \textit{Supervised Learning} & \multicolumn{12}{c}{\textit{ASR: 24.8\%}} \\
        & SSLBKD \cite{saha2022Backdoor} & 67.9\% & 30.1\% & 50.9\% & 80.3\% & 24.1\% & \underline{70.2}\% & 66.5\% & 29.1\% & \underline{51.2\%} & 70.9\% & 49.1\% & 33.9\% \\
        & CTRL \cite{li2023Embarrassingly} & 67.6\% & 67.6\% & 1.1\% & 76.3\% & 76.2\% & 4.7\% & 65.6\% & 65.4\% & 0.1\% & 69.2\% & 69.6\% & 0.1\% \\
        & CorruptEncoder \cite{zhang2024Data} & 68.0\% & 31.9\% & \underline{55.1\%} & 73.3\% & 40.1\% & 20.4\% & 66.1\% & 25.0\% & 26.1\% & 70.3\% & 39.1\% & \underline{42.1\%} \\
        & BLTO \cite{sun2023Backdoor} & 68.4\% & 35.5\% & 45.1\% & 72.1\% & 16.3\% & 77.6\% & 65.7\% & 44.2\% & 31.6\% & 70.1\% & 21.2\% & 51.0\% \\
        & BLTO-N  & 68.4\% & 35.5\% & 34.0\% & 72.1\% & 16.3\% & 47.1\% & 65.7\% & 44.2\% & 23.1\% & 70.1\% & 21.2\% & 33.8\% \\
        & \cellcolor{lightgray} Our NA & \cellcolor{lightgray} 68.3\% & \cellcolor{lightgray} 12.2\% & \cellcolor{lightgray} \textbf{84.8\%} & \cellcolor{lightgray} 79.2\% & \cellcolor{lightgray} 10.8\% & \cellcolor{lightgray} \textbf{71.4\%} & \cellcolor{lightgray} 66.5\% & \cellcolor{lightgray} 2.6\% & \cellcolor{lightgray} \textbf{97.1\%} & \cellcolor{lightgray} 70.1\% & \cellcolor{lightgray} 21.1\% & \cellcolor{lightgray} \textbf{64.8\%} \\
        \cmidrule(lr){2-14}
        & Oracle-Poisoning & 68.1\% & 2.4\% & 97.3\% & 79.0\% & 1.5\% & 98.5\% & 66.3\% & 3.5\% & 96.1\% & 70.3\% & 2.1\% & 97.7\% \\
        & BadEncoder \cite{jia2022BadEncoder} & \multicolumn{3}{c}{ASR: 97.1\%} & \multicolumn{3}{c}{ASR: 98.4\%} & \multicolumn{3}{c}{ASR: 94.2\%} & \multicolumn{3}{c}{ASR: 95.1\%} \\
        \midrule
        \multirow{9}{*}{CIFAR-10}
        & \textit{Supervised Learning} & \multicolumn{12}{c}{\textit{ASR: 80.9\%}} \\
        & SSLBKD \cite{saha2022Backdoor} & 82.0\% & 17.1\% & \underline{67.6\%} & 89.3\% & 48.2\% & \underline{40.1\%} & 70.1\% & 21.2\% & \underline{69.1\%} & 70.0\% & 18.2\% & \underline{69.2\%} \\
        & CTRL \cite{li2023Embarrassingly} & 82.3\% & 63.7\% & 11.2\% & 84.1\% &  80.2\% & 13.4\% & 72.9\% & 70.2\% & 13.4\% & 72.4\% & 60.0\% & 22.0\% \\
        & CTRL-NG & 79.0\% & 45.7\% & 40.1\% & 82.3\% & 39.7\% & 67.9\% & 70.4\% & 36.9\% & 68.5\% & 69.1\% & 12.3\% & 81.1\% \\
        & BLTO \cite{sun2023Backdoor} & 82.6\% & 10.9\% & 99.1\% & 81.3\% & 10.1\% & 99.4\% & 70.3\% & 11.0\% & 99.1\% & 71.1\% & 10.1\% & 98.7\% \\
        & BLTO-N  & 82.6\% & 10.9\% & 13.1\% & 81.3\% & 10.1\% & 9.1\% & 70.3\% & 11.0\% & 17.7\% & 71.1\% & 10.1\% & 15.6\% \\
        & \cellcolor{lightgray} Our NA & \cellcolor{lightgray} 82.8\% & \cellcolor{lightgray} 18.6\% & \cellcolor{lightgray} \textbf{75.9\%} & \cellcolor{lightgray} 89.9\% & \cellcolor{lightgray} 19.6\% & \cellcolor{lightgray} \textbf{72.6\%} & \cellcolor{lightgray} 70.8\% & \cellcolor{lightgray} 16.2\% & \cellcolor{lightgray} \textbf{79.9\%} & \cellcolor{lightgray} 68.3\% & \cellcolor{lightgray} 12.5\% & \cellcolor{lightgray} \textbf{85.6\%} \\
        \cmidrule(lr){2-14}
        & Oracle-Poisoning & 66.5\% & 15.2\% & 69.3\% & 89.5\% & 16.9\% & 74.1\% & 70.3\% & 18.1\% & 78.9\% & 68.9\% & 12.6\% & 79.5\% \\
        & BadEncoder \cite{jia2022BadEncoder} & \multicolumn{3}{c}{ASR: 72.2\%} & \multicolumn{3}{c}{ASR: 81.8\%} & \multicolumn{3}{c}{ASR: 65.1\%} & \multicolumn{3}{c}{ASR: 77.9\%} \\
        \bottomrule
    \end{tabular}
\label{tab:targeted_performance}
\end{table*}

\begin{figure*}
    \centering
    \includegraphics[width=\linewidth]{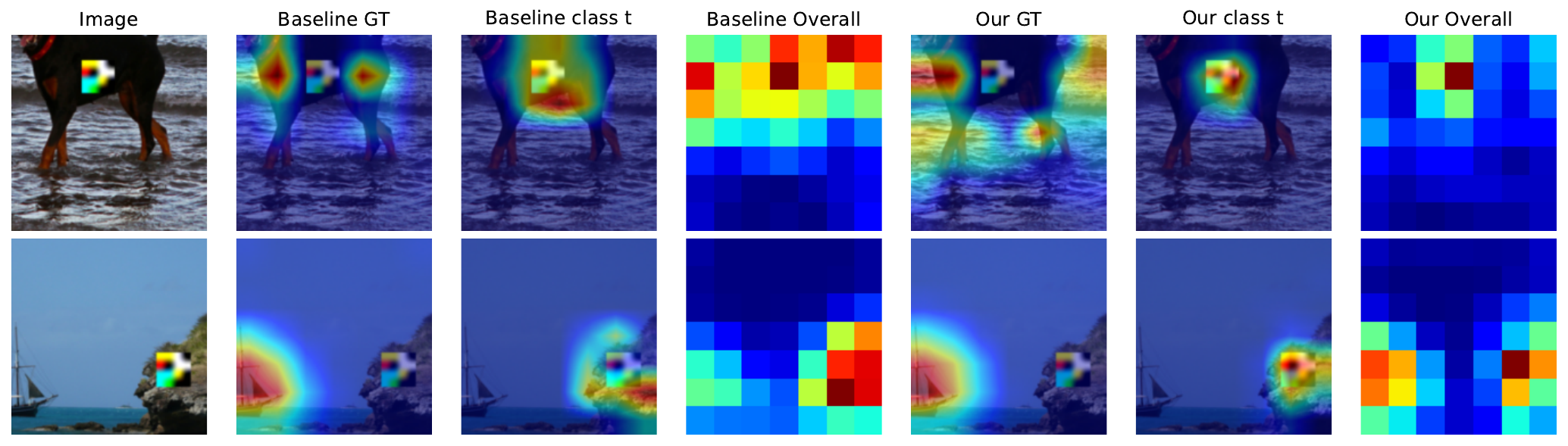}
    \caption{Class activation maps (CAM) \cite{selvaraju2017GradCAM} of attacks. GT means Ground Truth and class $t$ means attack target. Our attack produces a more focused heatmap.}
    \label{fig:heatmap}
\end{figure*}

\section{Experiments}
\subsection{Experimental Setup}
\textbf{Datasets.} We primarily use ImageNet-100 and CIFAR-10~\cite{krizhevsky2009learning} for evaluation. ImageNet-100 is a 100-class subset of ImageNet-1K~\cite{deng2009imagenet}, with the split provided by \cite{saha2022Backdoor}.
We randomly sample a 50K subset from CC3M~\cite{sharma2018Conceptual}, called CC-50K, to train the CLIP model which then is evaluated on ImageNet-1k.

\noindent\textbf{Evaluation.} We benchmark four contrastive learning frameworks: MoCo v2~\cite{chen2020Improved}, SimCLR~\cite{chen2020Simple}, BYOL~\cite{grill2020Bootstrap}, and SimSiam~\cite{chen2021Exploring}. Unless otherwise noted, we adopt MoCo v2 with a ResNet-18 backbone as the default pre-training setup and conduct all experiments on ImageNet-100. After pre-training, we freeze the encoder and train a linear classifier on top for downstream evaluation. Following the normalization trick in~\cite{he2022Masked}, we apply $\ell_2$ feature normalization to stabilize training.
We compare our attack with four state-of-the-art self-supervised backdoor baselines: SSLBKD~\cite{saha2022Backdoor}, CTRL~\cite{li2023Embarrassingly}, CorruptEncoder+~\cite{zhang2024Data}, and BLTO~\cite{sun2023Backdoor}. For CorruptEncoder we adopt their official reference images and report the results of the improved CorruptEncoder+. We report clean accuracy (CA), backdoored accuracy (BA), and attack success rate (ASR). Unless specified otherwise, all metrics are measured at convergence rather than at the best intermediate checkpoint.

\noindent\textbf{Attack Settings.}
Following former practice~\cite{saha2022Backdoor,zhang2024Data}, we inject $\sim $650 poisoned images for ImageNet-100 and $\sim $2500 poisoned images for CIFAR-10 (poisoning ratio 0.5\%). The shadow and reference sizes are set equal to the number of poisoned images. The triggers are from~\cite{saha2022Backdoor} and will be resized to 50$\times$50 for ImageNet-100 and 8$\times$8 for CIFAR-10. More details can be found in Appendix.

\subsection{Attack Effectiveness}
\label{sec:attack_effectiveness}

Table \ref{tab:targeted_performance} reports attack results on ImageNet-100 and CIFAR-10. Our method consistently delivers state-of-the-art ASR across all datasets and self-supervised methods, even surpassing the oracle BadEncoder on CIFAR-10 for MoCo v2, SimSiam, and SimCLR.
BLTO attains high ASR (consistently $\sim$99\% on CIFAR-10), though its poisons exhibit strong target-class semantics that yield 80-90\% ASR even without backdoor training. We therefore normalize ASR using clean encoder baselines. Both CTRL and BLTO rely on invisible triggers that are especially sensitive to CL augmentations, leading to a noticeable performance drop on the larger ImageNet-100. Removing Gaussian blur (a common CL augmentation) notably boosts CTRL ASR on CIFAR-10 as shown in Table \ref{tab:targeted_performance}.
For reference, we trained supervised models with CrossEntropy on the SSLBKD poisons. Supervised models achieve 24.8\% (80.9\%) ASR on ImageNet-100 (CIFAR-10).
All evaluated attacks maintain encoder utility, achieving performance comparable to clean encoders, shown in Appendix, across datasets and CL methods.

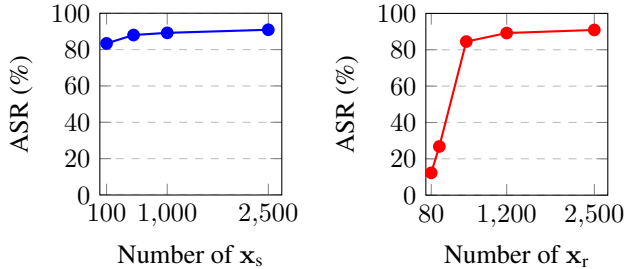
\begin{figure}[!t]
    \centering
    \begin{subfigure}[b]{0.48\columnwidth}
        \centering
        \begin{tikzpicture}
            \begin{axis}[
                width=\columnwidth,
                height=4cm,
                xlabel={Number of $\mathbf{x}_{\text{s}}$},
                ylabel={ASR (\%)},
                xmin=0, xmax=2700,
                ymin=0, ymax=100,
                xtick={100, 1000, 2500},
                ytick={0, 20, 40, 60, 80, 100},
                ymajorgrids=true,
                grid style=dashed,
                legend pos=south east,
                legend style={font=\small},
                ]
                \addplot[mark=*,thick,blue] coordinates {
                    (100, 83.4)
                    (500, 88.0)
                    (1000, 89.2)
                    (2500, 90.9)
                };
            \end{axis}
        \end{tikzpicture}
    \end{subfigure}
    \hfill
    \begin{subfigure}[b]{0.48\columnwidth}
        \centering
        \begin{tikzpicture}
            \begin{axis}[
                width=\columnwidth,
                height=4cm,
                xlabel={Number of $\mathbf{x}_{\text{r}}$},
                ylabel={ASR (\%)},
                xmin=0, xmax=2700,
                ymin=0, ymax=100,
                xtick={80, 1200, 2500},
                ytick={0, 20, 40, 60, 80, 100},
                ymajorgrids=true,
                grid style=dashed,
                legend pos=south east,
                legend style={font=\small},
                ]
                \addplot[mark=*,thick,red] coordinates {
                    (80, 12.3)
                    (200, 26.8)
                    (600, 84.5)
                    (1200, 89.2)
                    (2500, 90.9)
                };
            \end{axis}
        \end{tikzpicture}
    \end{subfigure}
    \caption{ Impact of (a) size of shadow images and (b) size of reference images.}
    \label{fig:reference_impact}
\end{figure}

\noindent\textbf{Multiple Target Classes.}
Table~\ref{tab:multi_target} summarizes the attack performance targeting several categories. Each class is assigned distinct trigger from \cite{saha2022Backdoor} while keeping the per-class poisoning ratio fixed at 50\%. We adopt SimSiam for pre-training.
As the number of attacked classes increases, ASR drops because the model capacity is shared among more objectives. Nonetheless, our proposed NA retains a strong ASR of 92.7\% even in the challenging four-class scenario, underscoring its scalability to multi-target settings.

\begin{table}[t]
    \centering
    \small
    \caption{ASR of NA on CLIP.}
    \label{tab:poisoning_CLIP}
    \begin{tabular}{cccc}
    \toprule
    \textbf{Image-text Pairs} & \textbf{Top1} & \textbf{Top5} \\
    \midrule
    Reference images + Reference texts & 87.1\% & 95.9\%  \\
    Noisy images + Reference texts & 100.0\% & 100.0\% \\
    \bottomrule
    \end{tabular}
\end{table}

\begin{table}[t]
    \centering
    \footnotesize
    \caption{Multi-target attack of NA.}
    \label{tab:multi_target}
    \begin{tabular}{lcc}
    \toprule
    \textbf{Class Names} & \textbf{CA} & \textbf{ASR} \\
    \midrule
    Shih-Tzu, Ski Mask & 66.2\% & 97.4\% \\
    Carbonara, Mixing Bowl & 66.4\% & 98.3\% \\
    Honeycomb, Little Blue Heron, Coyote & 65.7\% & 96.7\% \\
    Tripod, Ski Mask, Chesapeake Bay Retriever & 65.9\% & 96.5\% \\
    Pickup Truck, Chihuahua, Vacuum, Bookcase & 65.6\% & 94.3\% \\
    Throne, Pedestal, Pickup Truck, Borzoi & 65.9\% & 92.7\% \\
    \bottomrule
    \end{tabular}
\end{table}

\begin{table}[t]
    \centering
    \caption{Attack performance across different image-text contrastive models.}
    \label{tab:network_performance}
    \footnotesize
    \begin{tabular}{l|cc}
    \toprule
    Different pipelines & ACC (\%) & ASR (\%) \\
    \midrule
    clip-base-patch16-224 + finetune & 60.2 & 100.0 \\
    siglip-base-patch16-224 + finetune & 65.1 & 99.0 \\
    clip-vit-base-patch32 + train from scratch & 16.2 & 93.1 \\
    \bottomrule
    \end{tabular}
  \end{table}

\subsection{Abalation Study}
\label{sec:abalation_study}

\begin{figure*}[t]
    \centering
    \begin{subfigure}[b]{0.24\textwidth}
        \centering
        \includegraphics[width=\linewidth]{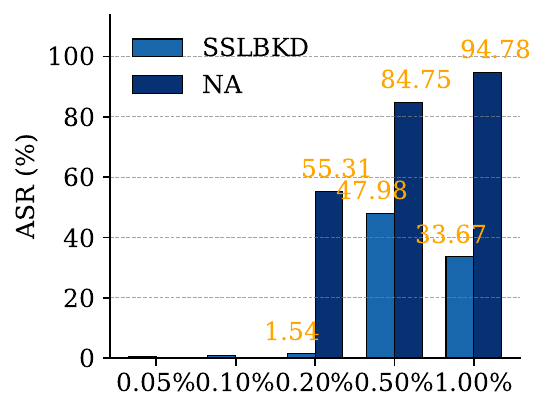}
        \caption{Poisoning ratio}
        \label{fig:poisoning_ratio}
    \end{subfigure}
    \begin{subfigure}[b]{0.24\textwidth}
        \centering
        \includegraphics[width=\linewidth]{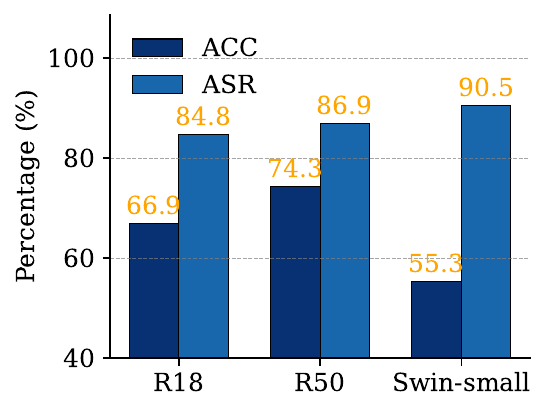}
        \caption{Architecture}
        \label{fig:architecture_comparison}
    \end{subfigure}
    \begin{subfigure}[b]{0.24\textwidth}
        \centering
        \includegraphics[width=\linewidth]{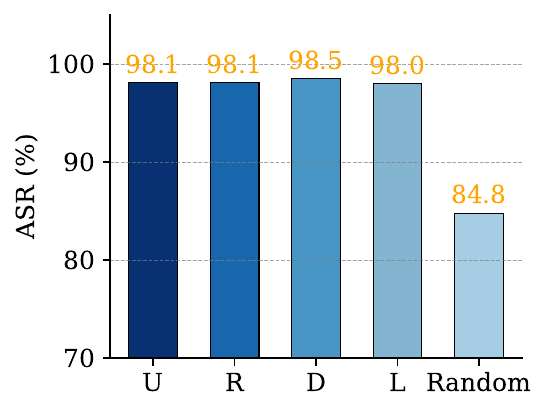}
        \caption{Layout}
        \label{fig:asr_vs_concatenation_direction}
    \end{subfigure}
    \begin{subfigure}[b]{0.24\textwidth}
        \centering
        \includegraphics[width=\linewidth]{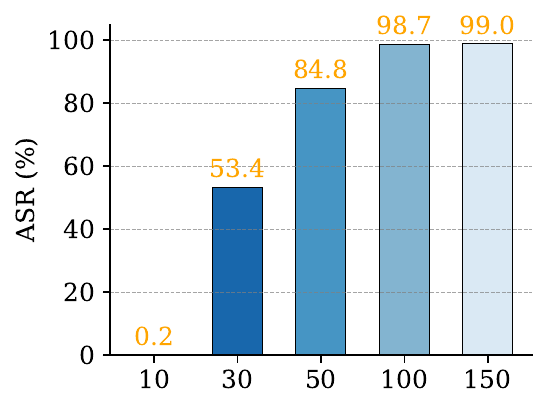}
        \caption{Trigger size}
        \label{fig:asr_vs_trigger_size}
    \end{subfigure}
    \caption{ASR with different settings.}
\end{figure*}

Figure \ref{fig:poisoning_ratio} shows the ASR of NA and SSLBKD with different poisoning ratios. Figure \ref{fig:architecture_comparison} shows the ASR of NA with different neural network architectures. Figure \ref{fig:asr_vs_trigger_size} shows the ASR with different trigger sizes. Our attack generalizes across architectures and achieves significant ASR (>50\%) at a poisoning ratio of 0.2\% and trigger size of 30$\times$30. Figure \ref{fig:asr_vs_concatenation_direction} evaluates ASR under four fixed layouts. Although a fixed layout achieves higher ASR, we adopt randomized layouts for better generalization. Figure \ref{fig:reference_impact} shows the impact of the number of shadow images and reference images on CIFAR10. We observe that the ASR saturates at around 200 shadow images and 1000 reference images, respectively.

\begin{table*}[t]
  \centering
  \caption{Detection performance of different detections.
  We mark the successful backdoor detection by marker * for DECREE, SSL-Cleanse, and Beatrix.
  For Beatrix, we mark images beyond 95-th percentile as poisoned images.}
  \label{tab:detection_performance}
  \setlength{\tabcolsep}{2pt}
  \resizebox{\textwidth}{!}{
  \begin{tabular}{l|ccc|ccc|ccc|ccc|ccc|ccc}
  \toprule
  & \multicolumn{12}{c|}{CIFAR10} & \multicolumn{6}{c}{ImageNet-100} \\
  \cmidrule{2-19}
  & \multicolumn{3}{c|}{DECREE \cite{feng2023Detecting}} & \multicolumn{3}{c|}{SSL-Cleanse \cite{zheng2024ssl}} & \multicolumn{3}{c|}{DeDe \cite{hou2025DeDe}} & \multicolumn{3}{c|}{Beatrix \cite{ma2023Beatrix}} & \multicolumn{3}{c|}{DECREE \cite{feng2023Detecting}} & \multicolumn{3}{c}{DeDe \cite{hou2025DeDe}} \\
  \cmidrule{2-19}
  Metric & BadEnc. & SSLBKD & Ours & BadEnc. & SSLBKD & Ours & BadEnc. & SSLBKD & Ours & BadEnc. & SSLBKD & Ours & BadEnc. & SSLBKD & Ours & BadEnc. & SSLBKD & Ours \\
  \midrule
  Recall & 0.99* & 0.92 & 0.92* & * & False & False & 0.81 & 0.61 & 0.73 & 0.87 & 0.66 & 0.96* & 0.82 & 0.82 & 0.99 & 0.69 & 0.71 & 0.49 \\
  Precision & 0.99* & 0.54 & 0.89* & * & False & False & 0.90 & 0.79 & 0.81 & 0.98 & 0.91 & 0.95* & 0.51 & 0.51 & 0.50 & 0.61 & 0.51 & 0.57 \\
  AUC & 1.0* & 0.47 & 0.96* & * & False & False & 0.93 & 0.82 & 0.87 & 0.97 & 0.91 & 0.99* & 0.52 & 0.52 & 0.31 & 0.67 & 0.52 & 0.58 \\
  \bottomrule
  \end{tabular}
  }
\end{table*}

\begin{table}[ht]
    \centering
    \small
    \caption{Performance under adaptive defenses.}
    \label{tab:defense_performance}
    \begin{tabular}{l|cc}
    \toprule
    Method & ACC (\%) & ASR (\%) \\
    \midrule
    baseline & 66.1 & 82.3 \\
    +minimal crop rario (0.8) & 42.9 & 0.5 \\
    +no random cropping & 36.1 & 0.9 \\
    \bottomrule
    \end{tabular}
  \end{table}

\section{Extension to Image-Text CL}
\label{sec:extension_to_image_text_CL}

Our framework naturally generalizes to the image-text contrastive setting. Consider a victim model that employs CLIP~\cite{radford2021Learning} to align images with their textual descriptions. We construct poisoned image-text pairs to maximize the cosine similarity between the embeddings of backdoored images and those of reference sentences that depict the target category (e.g., ``a photo of a dog''). In this formulation, the reference sentence assumes the same role as the reference image in the purely visual scenario.
To assess the attack, we train a CLIP model from scratch on the CC-50K dataset with CleanCLIP implementation~\cite{bansal2023CleanCLIP}. We randomly sample 250 clean images (merely $0.05\%$ of the training split) to craft noisy backdoored examples.
As reported in Table~\ref{tab:poisoning_CLIP}, our noisy alignment drives the ASR to 100\%. To mimic a practical scenario in which the defender is unaware of the underlying contrastive learning paradigm, we additionally inject various image-modal poisons directly into CLIP. The corresponding results are summarized in the Appendix Table 10.
Table~\ref{tab:network_performance} compares the attack effectiveness across various image-text models. By default, we fine-tune two pre-trained encoders, CLIP ViT-Base and SigLIP ViT-Base.
We further consider training CLIP ViT-Base from scratch for 50 epochs.
Despite reaching only 16.2\% top-1 accuracy, this scratch-trained model still attains a high ASR of 93.1\%.

\section{Defense}

We challenge our NA attack with commonly used defenses.

\begin{figure}[t]
    \centering
    \includegraphics[width=0.7\columnwidth]{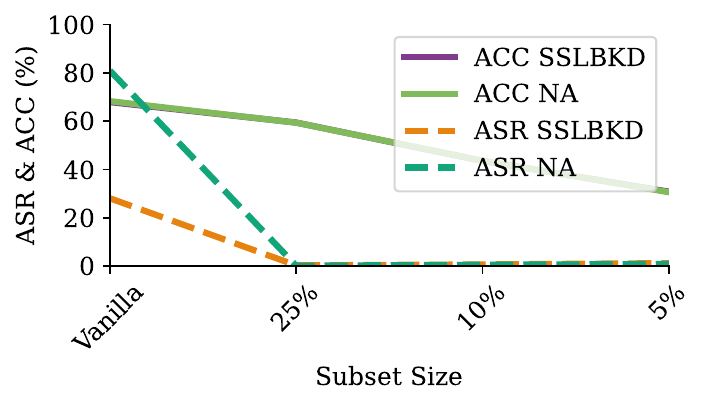}
    \caption{Distillation defense.}
    \label{fig:compress}
\end{figure}

\paragraph{Distillation.} We take the unsupervised distillation \textit{ComPress}~\cite{abbasikoohpayegani2020CompRess}. We adopt an available clean subset budget setup with 25\%, 10\%, and 5\%.  In Figure \ref{fig:compress}, we observed that unsupervised distillation effectively mitigates backdoor attacks, reducing the ASR to below 1\%. However, the cost is a significant reduction in the clean accuracy.

\paragraph{Detection.} We evaluate our attack against both supervised and self-supervised backdoor detection methods. We report the Recall (the proportion of detected poisons), Precision (the proportion of true poisons among detected poisons), and AUC (area under the ROC curve).
We employ the default threshold for DeDe~\cite{hou2025DeDe} and select the optimal one for DECREE~\cite{feng2023Detecting} via Youden's J statistic.
For supervised detection, we assume access to a 5\% clean data subset. While state-of-the-art methods can detect our attack on CIFAR-10, their performance degrades significantly on ImageNet-100. This indicates that attacks in high-dimensional spaces present considerable challenges to existing detections. More defenses can be found in Appendix.

\paragraph{Adaptive Defense.} We present the performance of adaptive defenses in Table \ref{tab:defense_performance}. NA relies on malicious co-occurrence from random augmentation, and the adaptive defense disrupts it. Although adaptive defense can effectively defend against NA, it may also impair the model's performance.

%% file: sec/supplementary.tex
\input{sec/proof.tex}

\begin{table}[t]
    \centering
    \footnotesize
    \caption{Clean performance on 10\% clean available subset.}
    \begin{tabular}{c|cc|cc|cc}
        \toprule
        \textbf{Dataset} & \multicolumn{2}{c|}{\textbf{MoCo v2}} & \multicolumn{2}{c|}{\textbf{BYOL}} & \multicolumn{2}{c}{\textbf{SimSiam}} \\
         & \textbf{ACC} & \textbf{ASR} & \textbf{ACC} & \textbf{ASR} & \textbf{ACC} & \textbf{ASR}  \\
        \midrule
        CIFAR10 & 69.0\% & 8.0\% & 88.3\% & 8.0\% & 71.1\% & 9.1\% \\
        ImageNet-100 & 66.5\% & 0.9\% & 80.1\% & 2.2\% & 66.1\% & 1.2\% \\
        \bottomrule
    \end{tabular}
    \label{tab:targeted_performance_clean}
\end{table}

\begin{figure*}[htbp]
    \centering
    \begin{subfigure}[b]{0.4\textwidth}
        \centering
        \includegraphics[width=\textwidth]{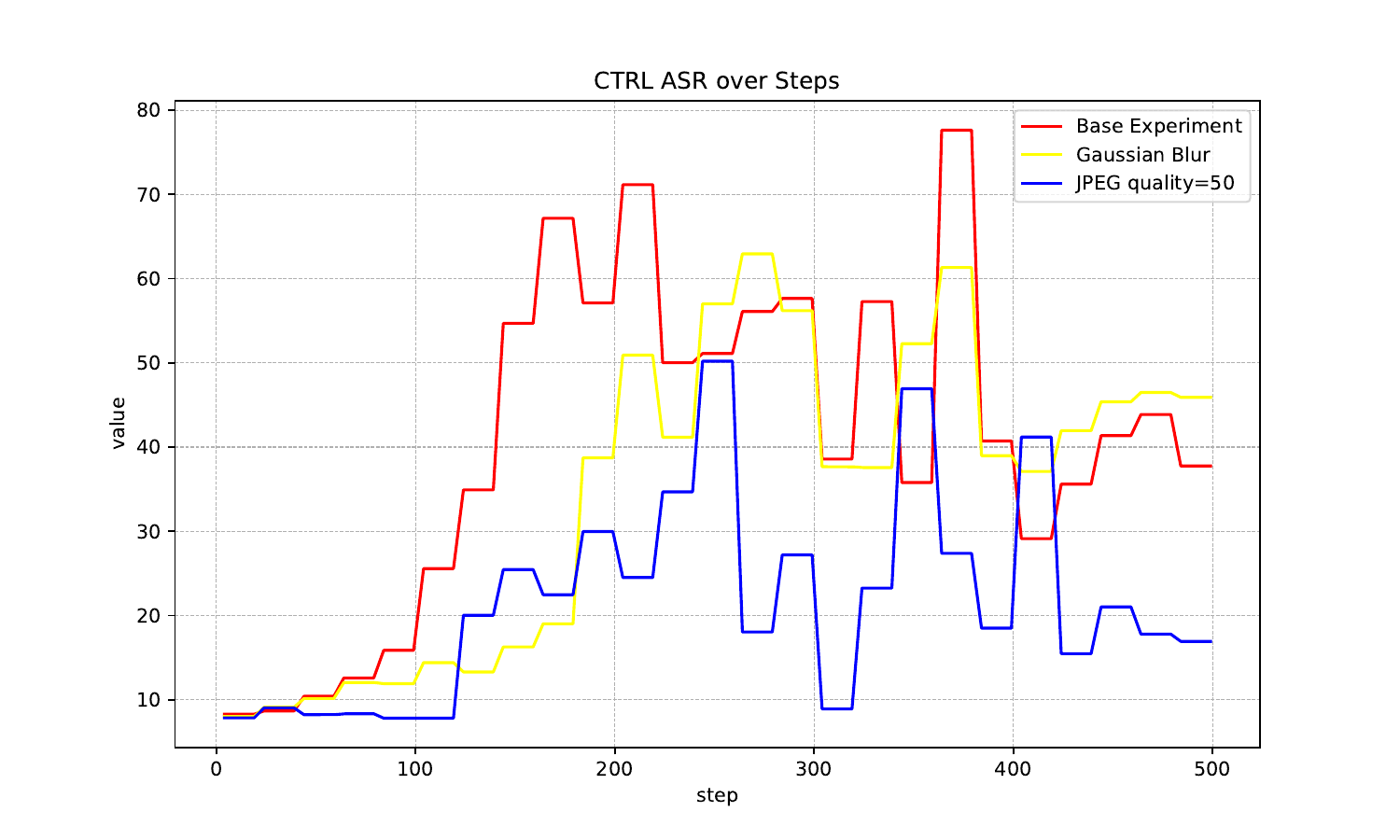}
        \caption{airplane}
        \label{fig:ctrl_class0}
    \end{subfigure}
    \hfill
    \begin{subfigure}[b]{0.4\textwidth}
        \centering
        \includegraphics[width=\textwidth]{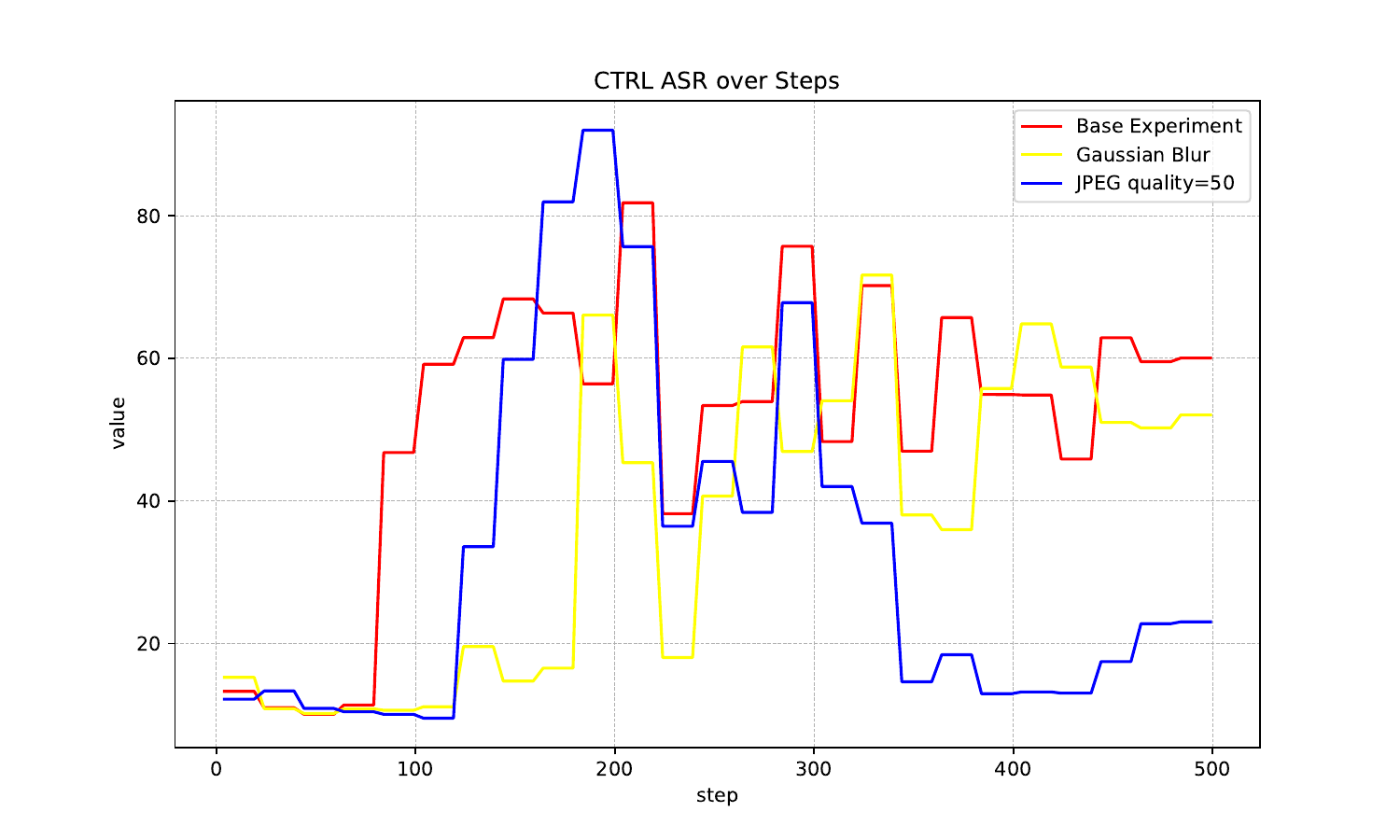}
        \caption{automobile}
        \label{fig:ctrl_class1}
    \end{subfigure}
    \begin{subfigure}[b]{0.4\textwidth}
        \centering
        \includegraphics[width=\textwidth]{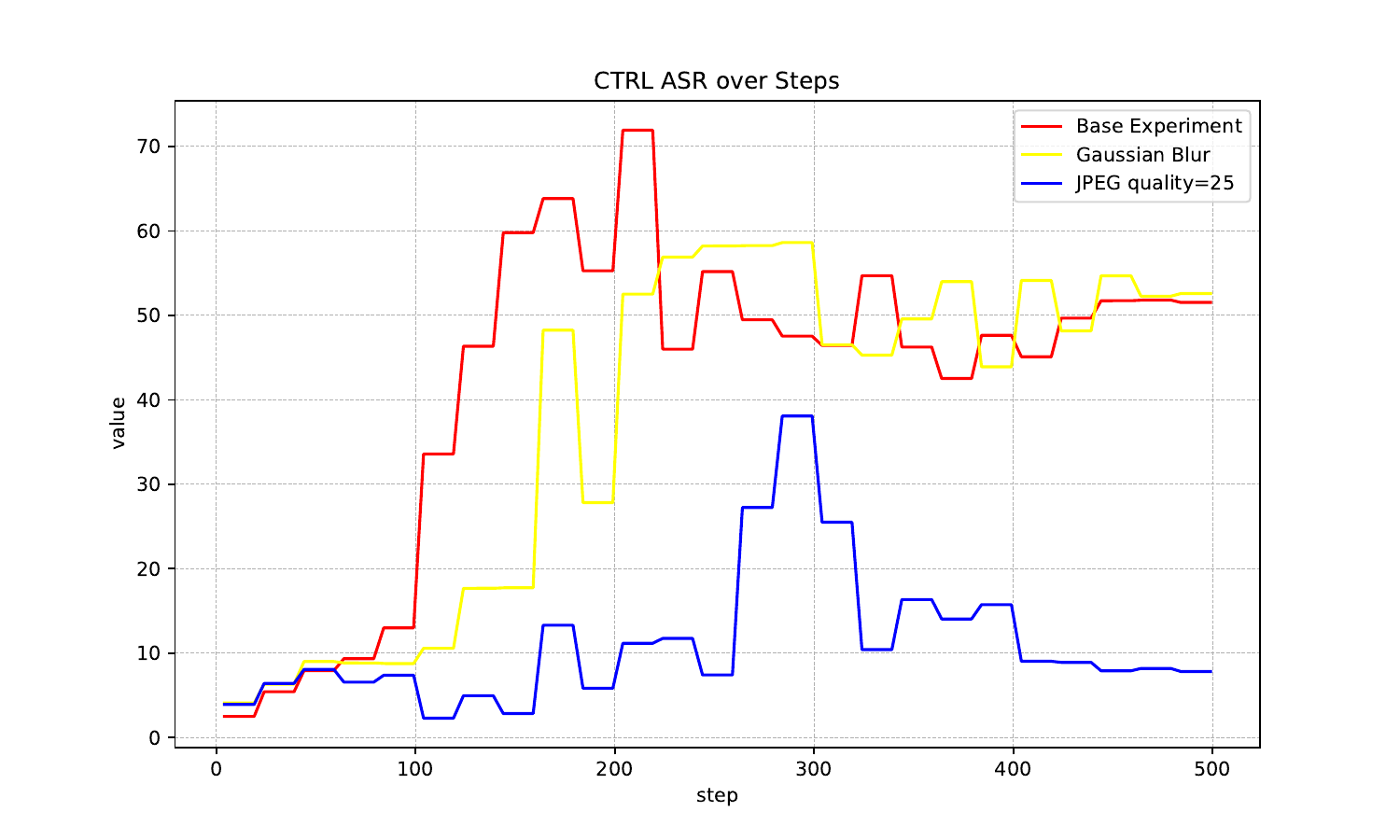}
        \caption{bird}
        \label{fig:ctrl_class2}
    \end{subfigure}
    \hfill
    \begin{subfigure}[b]{0.4\textwidth}
        \centering
        \includegraphics[width=\textwidth]{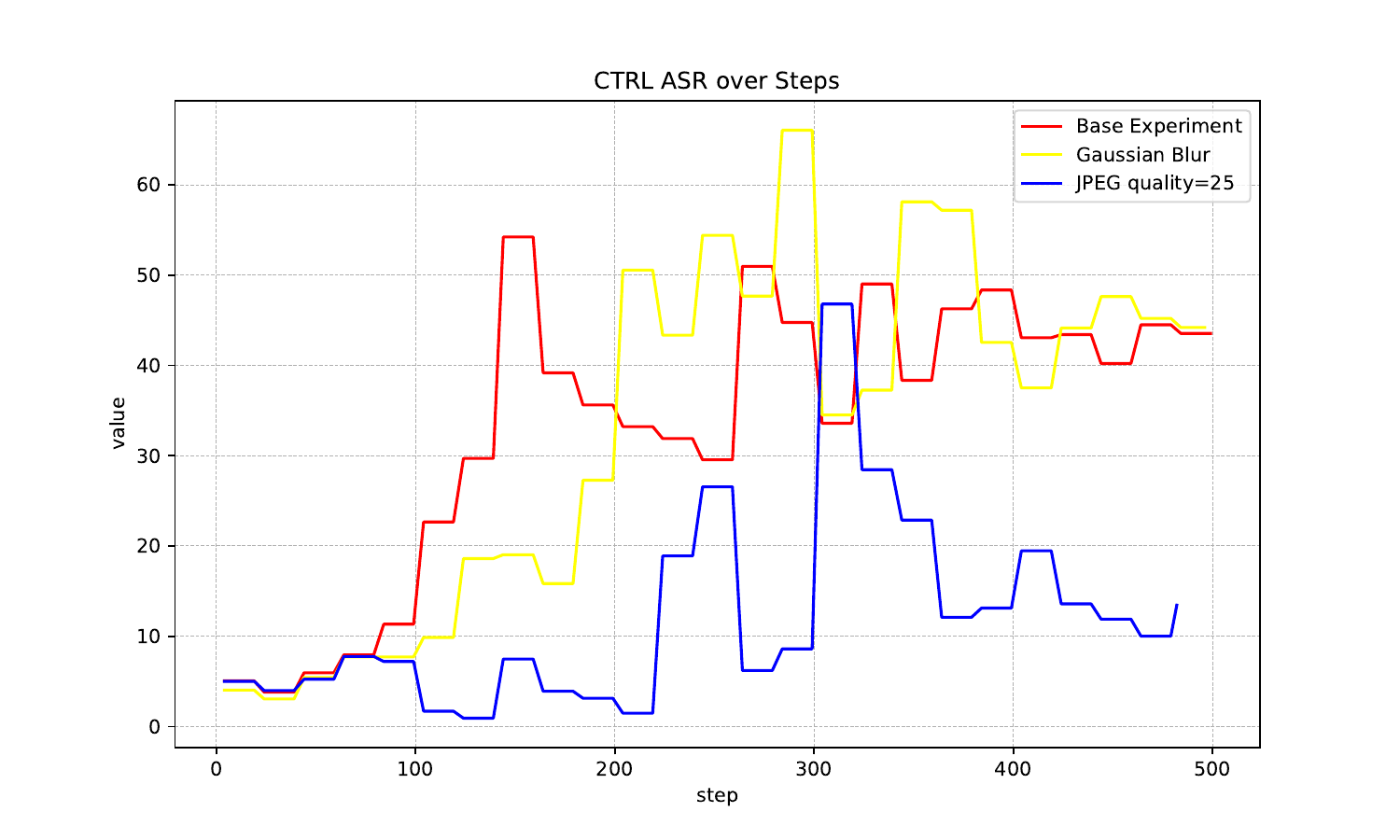}
        \caption{cat}
        \label{fig:ctrl_class3}
    \end{subfigure}
    \caption{Different attack classes of CTRL \cite{li2023Embarrassingly} on CIFAR-10 under various data processing methods. We use Gaussian noise and JPEG compression to perturb the poisons.}
    \label{fig:ctrl}
\end{figure*}

\begin{figure*}[htbp]
    \centering
    \begin{subfigure}[b]{0.49\linewidth}
        \includegraphics[width=\linewidth]{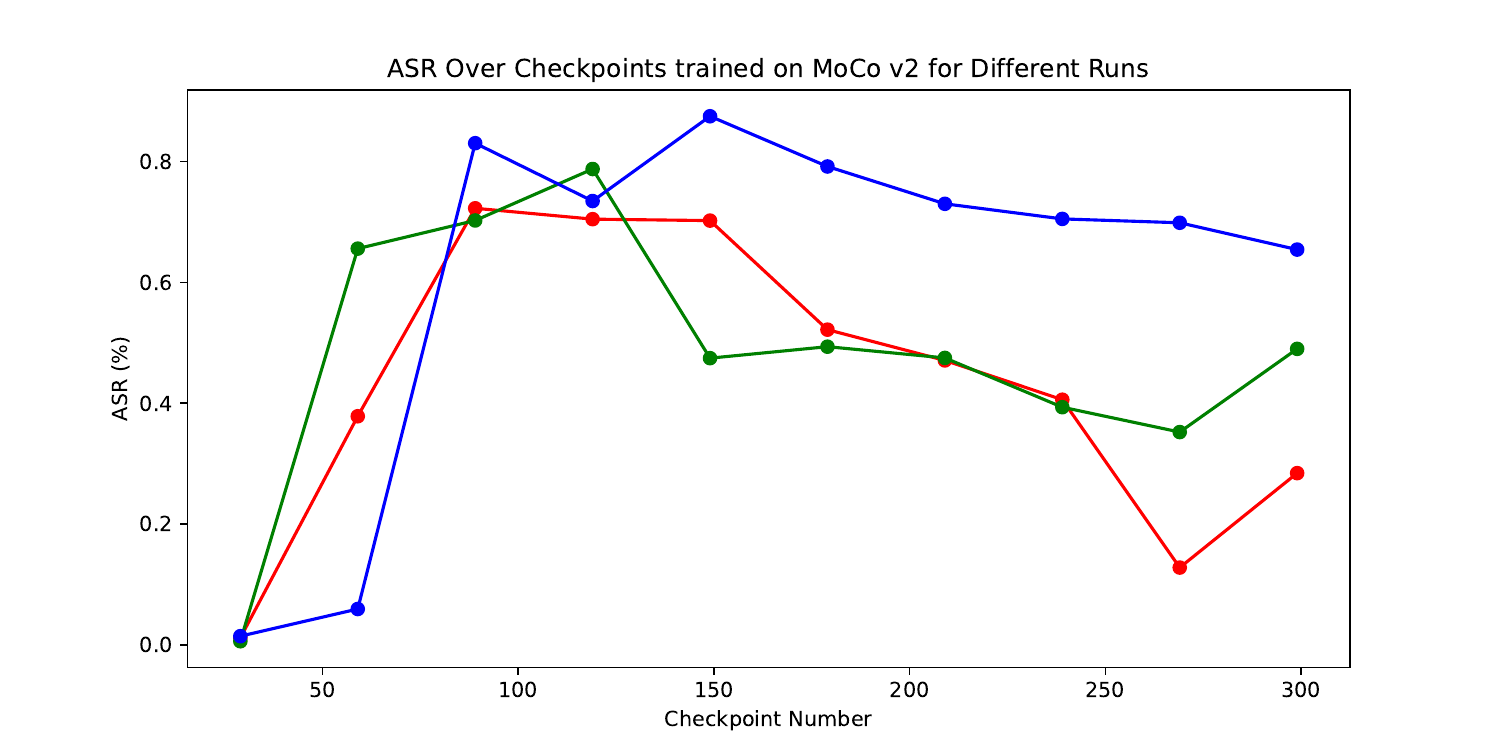}
    \end{subfigure}
    \hfill
    \begin{subfigure}[b]{0.49\linewidth}
        \includegraphics[width=\linewidth]{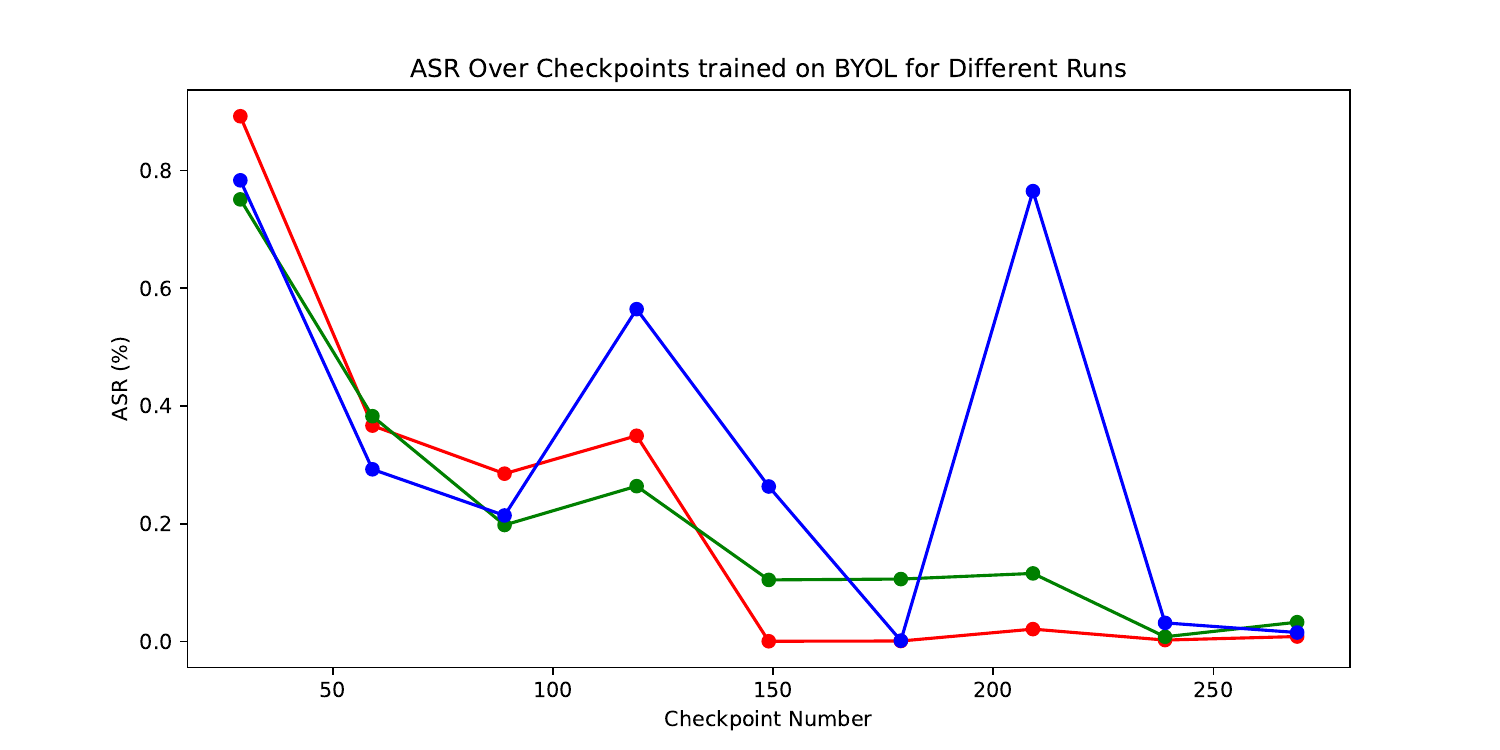}
    \end{subfigure}
    \caption{ASR over checkpoints of three CorruptEncoder \cite{zhang2024Data} trials on ImageNet-100.}
    \label{fig:appendix_asr_over_checkpoints}
\end{figure*}

\input{sec/sub_secs/information_theory}

\section{Experimental Details}

\begin{figure}[ht]
    \centering
    \begin{minipage}{\linewidth}
        \centering
        \includegraphics[width=0.3\linewidth]{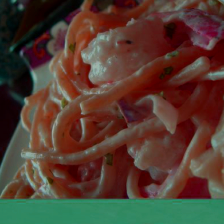}\hfill
        \includegraphics[width=0.3\linewidth]{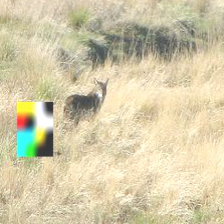}\hfill
        \includegraphics[width=0.3\linewidth]{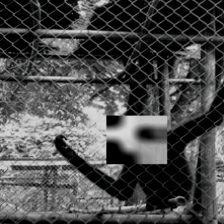}
        \newline
        \includegraphics[width=0.3\linewidth]{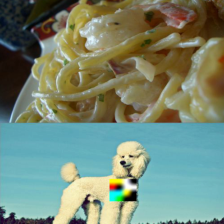}\hfill
        \includegraphics[width=0.3\linewidth]{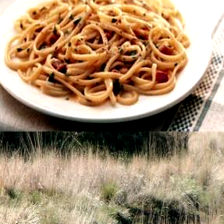}\hfill
        \includegraphics[width=0.3\linewidth]{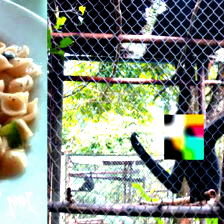}
    \end{minipage}
    \caption{\textbf{Augmented views of the poisoned data.} Each of the top row and the bottom row is one of the augmented views from the identical poisoned image of MoCo v2~\protect\cite{chen2020Improved} and the target attack class is carbonara.}
    \label{fig:augmented_images}
\end{figure}
\begin{figure}[!htbp]
    \centering

    \begin{subfigure}{0.16\linewidth}
        \centering
        \includegraphics[width=0.9\linewidth]{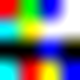}
        \caption{10}
    \end{subfigure}\hfill
    \begin{subfigure}{0.16\linewidth}
        \centering
        \includegraphics[width=0.9\linewidth]{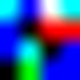}
        \caption{11}
    \end{subfigure}\hfill
    \begin{subfigure}{0.16\linewidth}
        \centering
        \includegraphics[width=0.9\linewidth]{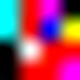}
        \caption{12}
    \end{subfigure}\hfill
    \begin{subfigure}{0.16\linewidth}
        \centering
        \includegraphics[width=0.9\linewidth]{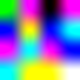}
        \caption{13}
    \end{subfigure}\hfill
    \begin{subfigure}{0.16\linewidth}
        \centering
        \includegraphics[width=0.9\linewidth]{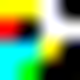}
        \caption{14}
    \end{subfigure}

    \vspace{2mm}

    \begin{subfigure}{0.16\linewidth}
        \centering
        \includegraphics[width=0.9\linewidth]{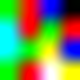}
        \caption{15}
    \end{subfigure}\hfill
    \begin{subfigure}{0.16\linewidth}
        \centering
        \includegraphics[width=0.9\linewidth]{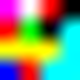}
        \caption{16}
    \end{subfigure}\hfill
    \begin{subfigure}{0.16\linewidth}
        \centering
        \includegraphics[width=0.9\linewidth]{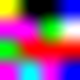}
        \caption{17}
    \end{subfigure}\hfill
    \begin{subfigure}{0.16\linewidth}
        \centering
        \includegraphics[width=0.9\linewidth]{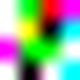}
        \caption{18}
    \end{subfigure}\hfill
    \begin{subfigure}{0.16\linewidth}
        \centering
        \includegraphics[width=0.9\linewidth]{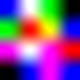}
        \caption{19}
    \end{subfigure}

    \caption{Illustration of the patch triggers.}
    \label{fig:triggers}
\end{figure}

\begin{table}[t]
  \centering
  \caption{Performance of irregular and invisible triggers.}
  \label{tab:trigger_strategies}
  \begin{tabular}{l|cc}
  \toprule
  Method & ACC (\%) & ASR (\%) \\
  \midrule
  baseline & 66.1 & 82.3 \\
  +Blended triggers \cite{chen2017Targeted} & 65.8 & 88.2 \\

  \bottomrule
  \end{tabular}
  \renewcommand{\arraystretch}{1.0}
\end{table}

\paragraph{Trigger.} We mainly use the trigger from \cite{saha2022Backdoor}, which are small square colorful patches, i.e. random 4$\times$4 RGB images, as Figure \ref{fig:triggers} shows. They are resized to the desired size when attached to the poisoned image. We demonstrate augmented views on ImageNet-100 in Figure \ref{fig:augmented_images}. We also use non-patch-based triggers to test our attacks, as shown in Table \ref{tab:trigger_strategies}.

\paragraph{Hyperparameters.} We synchronize the hyperparameters with the baseline SSLBKD~\cite{saha2022Backdoor}, ensuring the comparability. Note that we slightly scale the training length to 300 epochs, as SSL methods typically require longer to converge. We provide the pre-training configurations and linear probing configurations in Table \ref{tab:pre-training_config} and Table \ref{tab:linear_probing_config} respectively.

\begin{table*}[ht]
    \centering
    \begin{tabular}{l|ccc}
      \hline
      Methods & MoCo v2 \& SimSiam \& SimCLR & BYOL \\
      \hline
      Training Epochs & 40 & 100 \\
      Batch Size & 256 & 256  \\
      Optimizer & SGD & Adam  \\
      Learning Rate Schedule & MultiStepLR & ExponentialLR  \\
      Learning Rate & 0.01 & 0.01  \\
      Weight Decay & \(1 \times 10^{-4}\) & \(5 \times 10^{-6}\)  \\
      Momentum & 0.9 & - \\
      \hline
      Resize \& Crop & RandomResizeAndCrop & RandomResizeAndCrop  \\
      RandomHorizontalFlip & 0.5 & 0.5  \\
      \hline
      \end{tabular}
    \caption{Hyperparameters for linear probing.}
    \label{tab:linear_probing_config}
  \end{table*}

\begin{figure}[t]
    \centering
    \includegraphics[width=\linewidth]{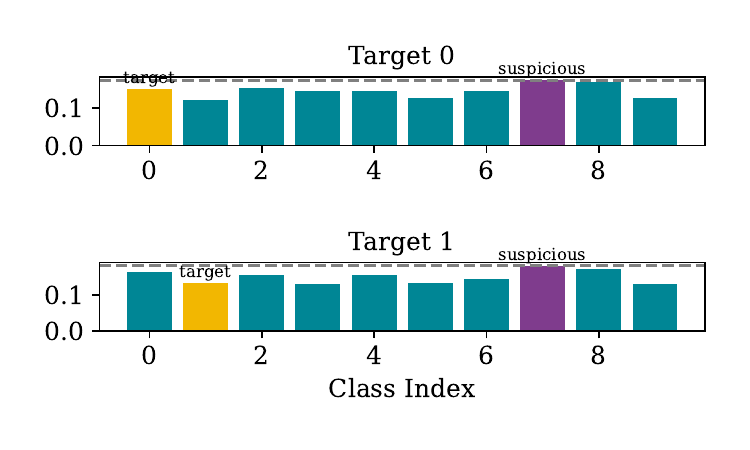}
    \vspace{-0.3in}
    \caption{Activation Cluster defense.}
    \label{fig:activation_cluster}
\end{figure}

\begin{table}[t]
    \centering
    \small
    \caption{ASR of directly poisoning CLIP with different image-modal poisons.}
    \label{tab:poisoning_CLIP_with_image_modal_poisons}
    \begin{tabular}{ccccc}
    \toprule
    \textbf{Metrics} & \textbf{SSLBKD} & \textbf{SIG} & \textbf{Gaussian noise} & \textbf{NA} \\
    \midrule
    Top1 & 99.9\% & 59.3\% & 99.8\% & 91.3\% \\
    Top5 & 99.9\% & 63.3\% & 99.9\% & 96.0\% \\
    \bottomrule
    \end{tabular}
\end{table}

\begin{table}[t]
    \centering
    \caption{SCAn results on CIFAR10 and ImageNet-100.}
    \small
    \begin{tabular}{ccccc}
    \toprule
    \textbf{Dataset} & \multicolumn{2}{c}{\textbf{CIFAR10}} & \multicolumn{2}{c}{\textbf{ImageNet-100}} \\
    \cmidrule(lr){2-3} \cmidrule(lr){4-5}
     & \textbf{MoCo v2} & \textbf{SimSiam} & \textbf{MoCo v2} & \textbf{SimSiam} \\
    \midrule
    CAP & 100\% & 100\% & 0\% & 0\% \\
    TPR & 11.5\% & 28.7\% & 26.2\% & 3.7\% \\
    FPR & 0.0\% & 0.1\% & 3.0\% & 4.9\% \\
    \bottomrule
    \end{tabular}
    \label{tab:scan}
\end{table}

\begin{table}[t]
    \centering
    \caption{PatchSearch defense.}
    \label{tab:patchsearch}
    \begin{tabular}{lcc}
        \toprule
        \textbf{Metric} & \textbf{MoCo} & \textbf{SimSiam} \\
        \midrule
        Poisons Removed & 38,710 & 28,666 \\
        Recall (\%) & 46.3 & 49.1 \\
        Precision (\%) & 0.8 & 1.2 \\
        ASR after defense (\%) & 61.0 & 77.1 \\
        \bottomrule
    \end{tabular}
\end{table}

\section{More Analysis of Attack Dynamics}
\label{sec:asr_analysis}

\begin{figure*}[t]
    \centering
    \begin{subfigure}[b]{0.45\textwidth}
        \includegraphics[width=\textwidth]{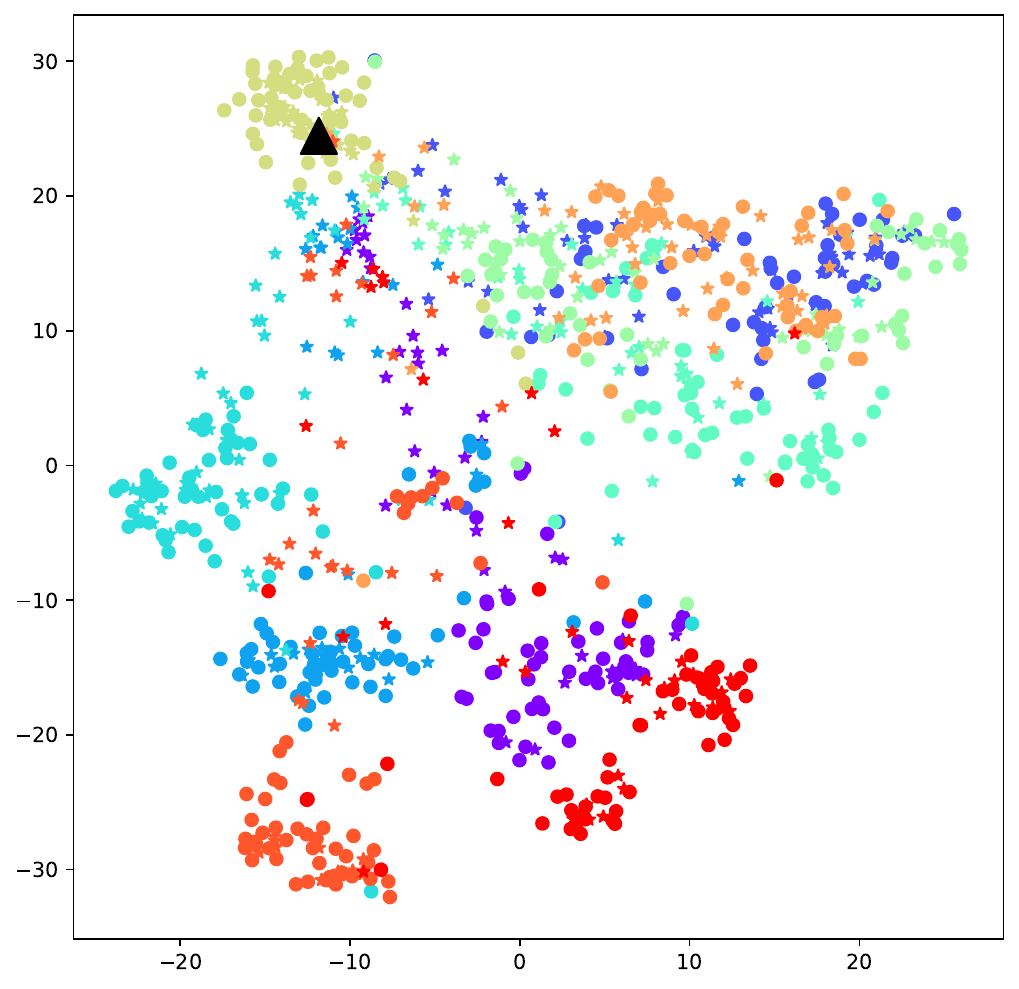}
    \caption{MoCo v2}
    \end{subfigure}
    \hfill
    \begin{subfigure}[b]{0.45\textwidth}
        \includegraphics[width=\textwidth]{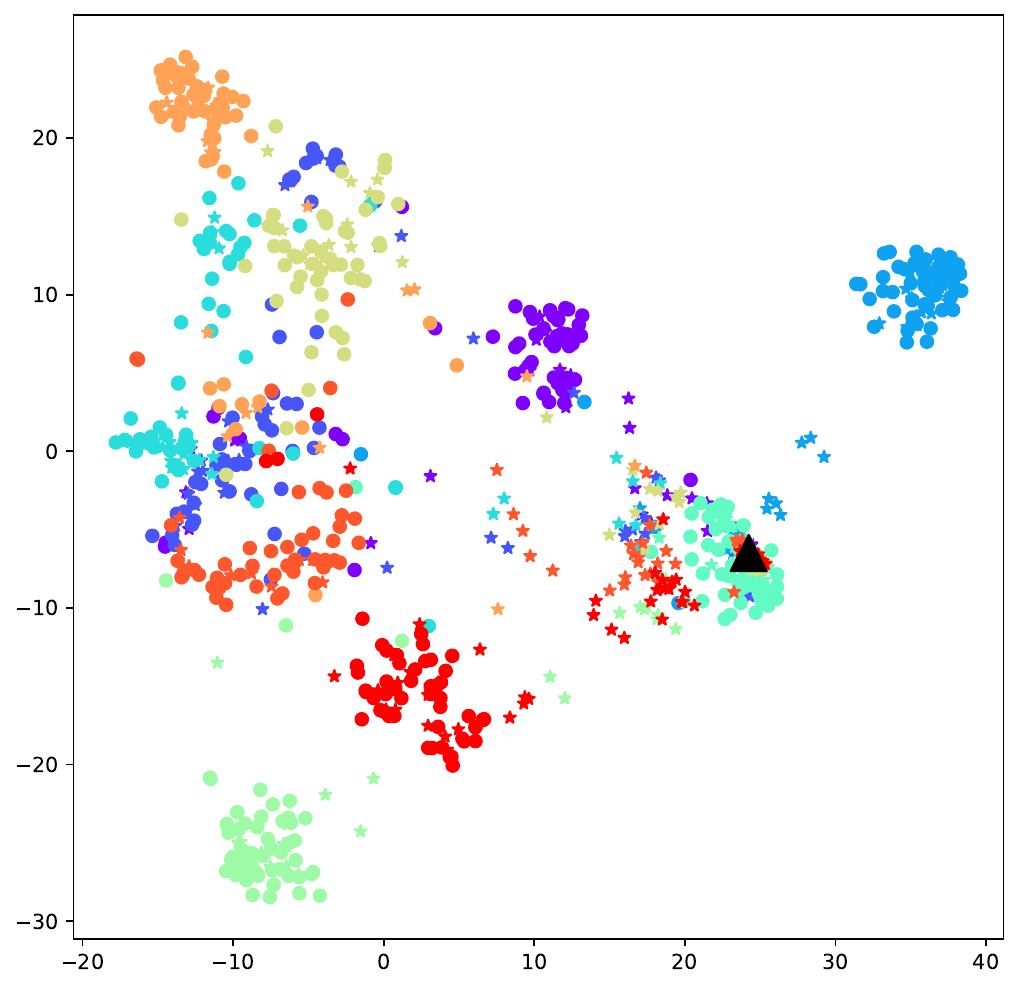}
    \caption{SimCLR}
    \end{subfigure}
    \par\medskip
    \begin{subfigure}[b]{0.45\textwidth}
        \includegraphics[width=\textwidth]{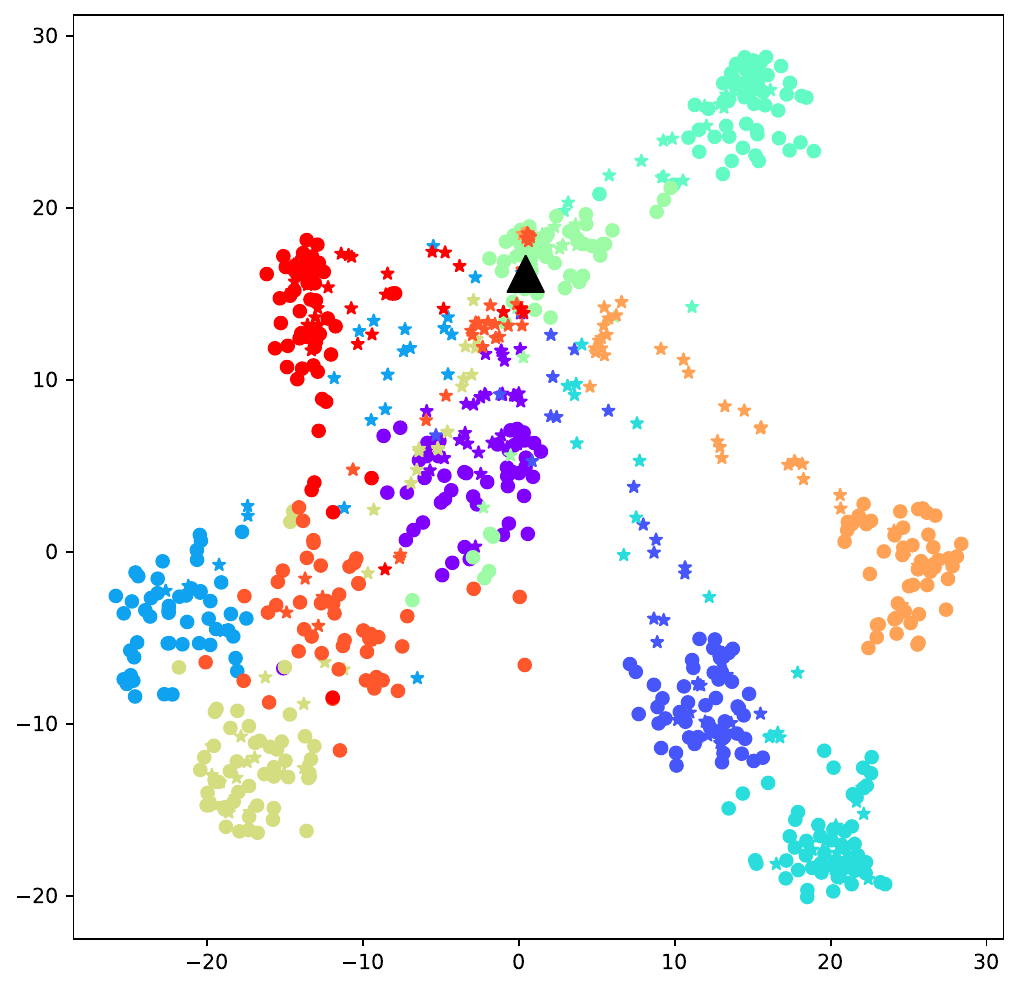}
    \caption{BYOL}
    \end{subfigure}
    \hfill
    \begin{subfigure}[b]{0.45\textwidth}
        \includegraphics[width=\textwidth]{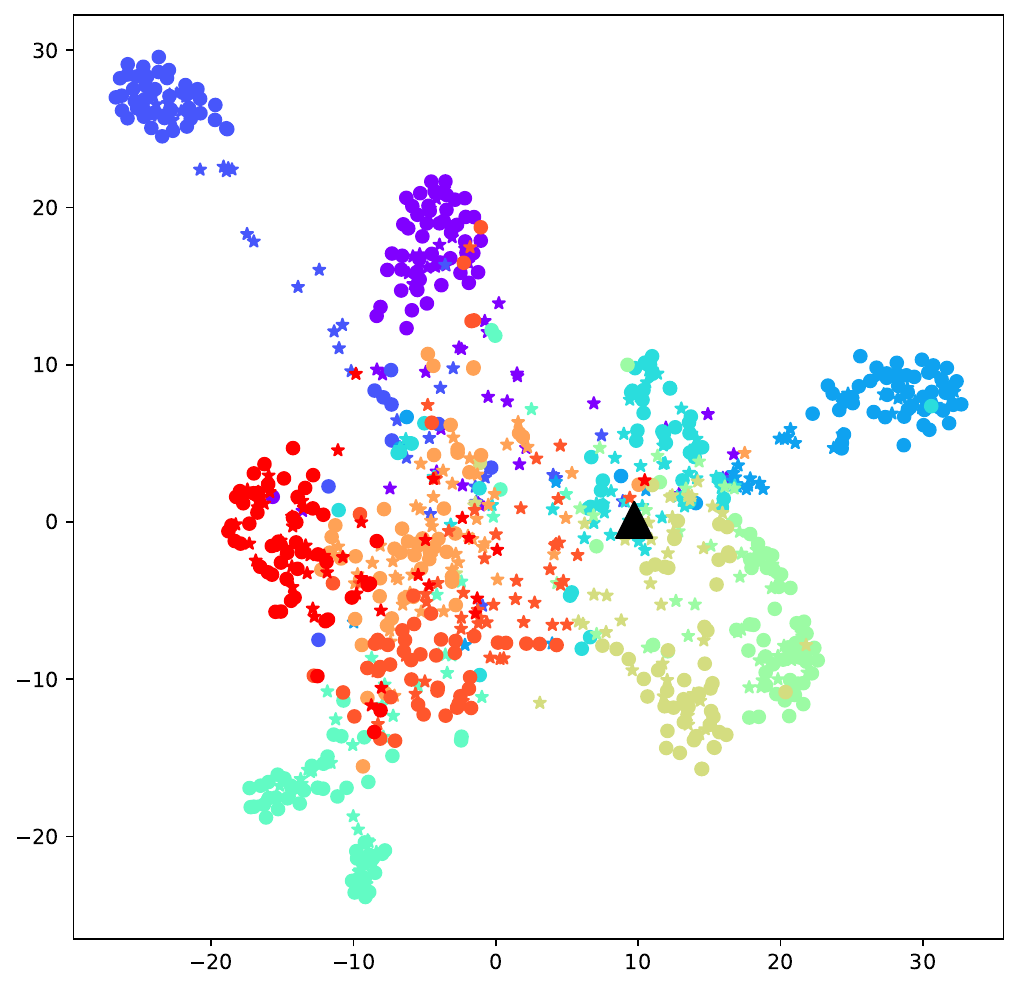}
    \caption{SimSiam}
    \end{subfigure}
    \caption{t-SNE visualization of the representation space of our attack. Black triangles \ding{115} are poison centers and colors represent different classes. Star and circle markers represent the poisoned and clean samples, respectively.}
    \label{fig:overall}
\end{figure*}

\paragraph{Decline in attack performance during the late training stage.}
In Figure~\ref{fig:appendix_asr_over_checkpoints} we plot the ASR trajectory of CorruptEncoder on ImageNet\,-\,100. The attack converges swiftly, attaining 60--80\% ASR within the first 50--100 epochs for both MoCo~v2 and BYOL. Training beyond this point, however, often causes the ASR to degrade. We conjecture that the Uniformity regularization in later epochs~\cite{wang2020Understanding} loosens the coupling between the backdoor and its reference image, echoing the observations of Sun\,\textit{et al.}~\cite{sun2023Backdoor}. A comparable trend is also visible in CTRL~\cite{li2023Embarrassingly} (Figure~\ref{fig:ctrl}), underscoring the generality of this phenomenon.

\begin{figure*}[t]
    \centering
    \begin{subfigure}[b]{\textwidth}
        \centering
        \includegraphics[width=\textwidth]{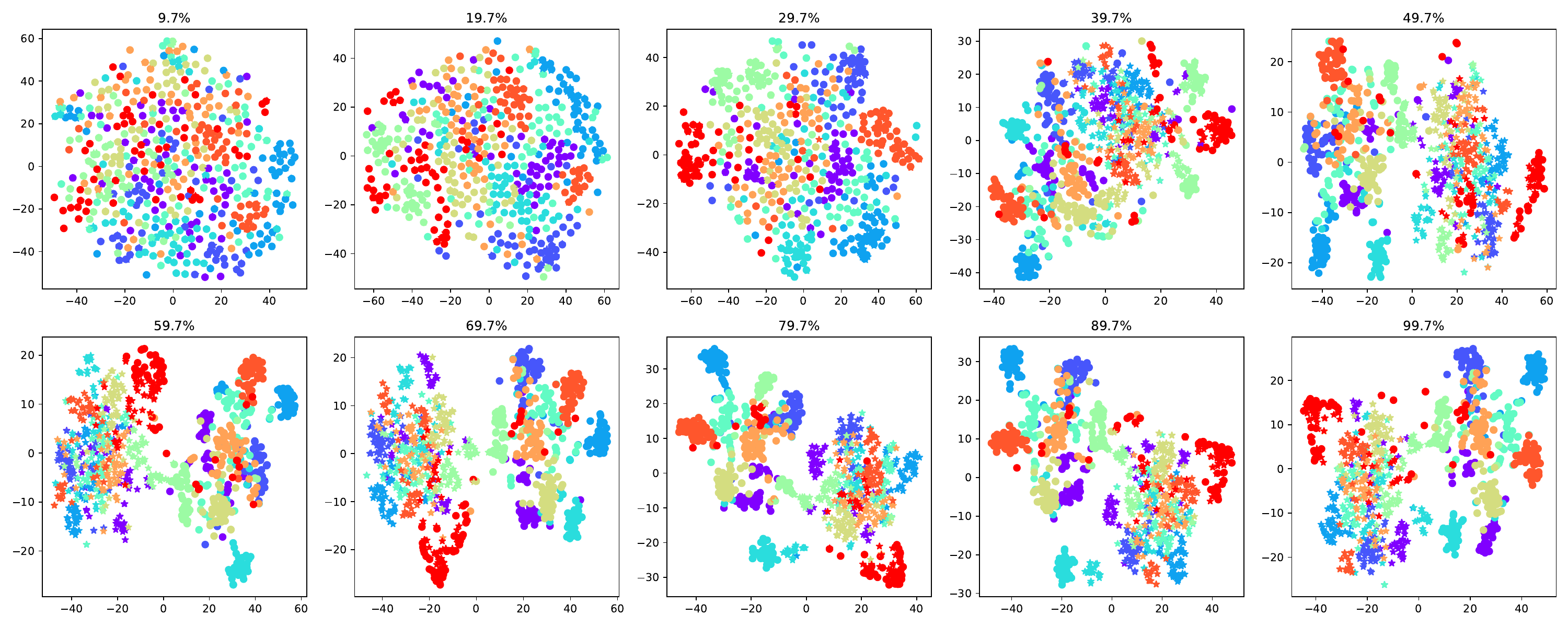}
        \caption{Attack category n03085013.}
    \end{subfigure}
    \begin{subfigure}[b]{\textwidth}
        \centering
        \includegraphics[width=\textwidth]{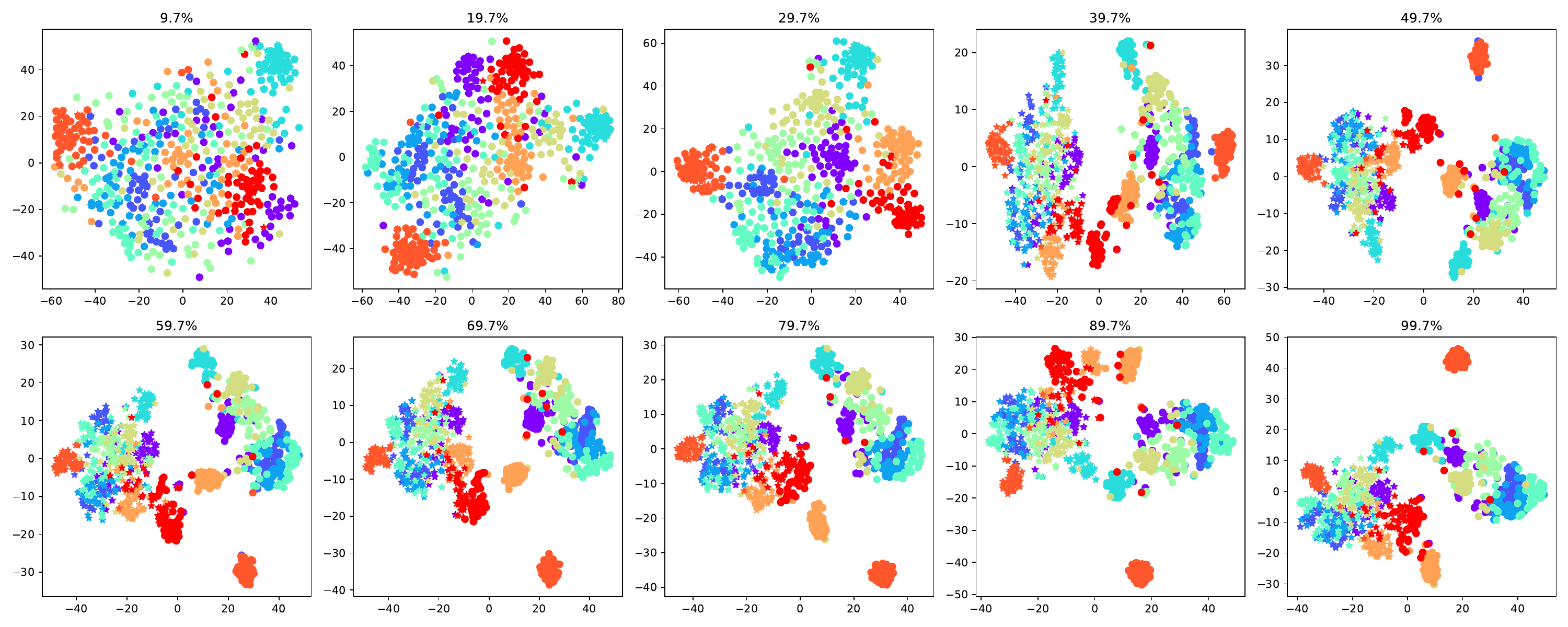}
        \caption{Attack category n03947888.}
    \end{subfigure}
    \caption{t-SNE visualization at various training stages on ImageNet-100. Circles represent clean samples, while stars denote poisons. Different classes are distinguished by color.}
    \label{fig:na_t-SNE}
\end{figure*}

\paragraph{Representation Visualization.}
Figure~\ref{fig:na_t-SNE} shows intermediate t\,-\,SNE snapshots, while Figure~\ref{fig:overall} depicts the representation space at convergence. Figure \ref{fig:na_t-SNE} shows that our attack can maintain the separability of poison representations in the later stages of training.

\begin{table}[t]
    \centering
    \footnotesize
    \caption{ASR on difference reference distributions.}
    \label{tab:downstream_distribution}
    \begin{tabular}{c|c|c|cc}
    \toprule
    \textbf{Pre-training} & \textbf{Reference} & \textbf{Model} & \multicolumn{2}{c}{\textbf{Results}} \\
    \textbf{Dataset} & \textbf{Dataset} & & \textbf{CA} & \textbf{ASR} \\
    \midrule
    \multirow{6}{*}{ImageNet-100} & \multirow{2}{*}{ImageNet-100-O} & MoCo v2 & 61.1\% & 77.1\% \\
    & & SimSiam & 54.7\% & 84.3\% \\
    \cmidrule(lr){2-5}
    & \multirow{2}{*}{STL-10} & MoCo v2 & 70.2\% & 59.0\% \\
    & & SimSiam & 70.5\% & 52.1\% \\
    \cmidrule(lr){2-5}
    & \multirow{2}{*}{CIFAR-10} & MoCo v2 & 52.2\% & 42.9\% \\
    & & SimSiam & 53.5\% & 49.8\% \\
    \bottomrule
    \end{tabular}
\end{table}

\paragraph{Reference Distribution Shift.}
Table~\ref{tab:downstream_distribution} investigates the attack effectiveness under a distribution mismatch between the pre-training and downstream. ImageNet-100-O is an alternative subset that is disjoint from ImageNet-100.
Such a shift hampers both benign performance and attack strength, since feature representations become sub-optimal for the new domain. Nevertheless, NA still delivers competitive attack efficacy, demonstrating that it can effectively generalize beyond the original pre-training distribution.

\section{More Defenses}

\noindent\textbf{PatchSearch}. \emph{PatchSearch} \cite{tejankar2023Defending} is a poison detection method design for SSL. Table \ref{tab:patchsearch} shows PatchSearch retrieves about half of the poisons, but the ASR remains high (61.0\% for MoCo v2 and 77.1\% for SimSiam).

\paragraph{Statistical Contamination Analyzer (SCAn)}.
We evaluated the \textit{SCAn} using three metrics: accuracy of the poisoned class prediction (CAP), false positive rate (FPR), and true positive rate (TPR). We implemented SCAn on CIFAR10 following \cite{li2023BackdoorBox} and randomly sampled 10\% of the test set to build the decomposition model. Table \ref{tab:scan} shows SCAn can effectively identify the poisoned class on CIFAR10, yet it is entirely out of work on the larger ImageNet-100.

\paragraph{Activation Clustering (AC)}.
The \textit{AC} \cite{chen2018Detecting} detection is based on the intuition that poisoned examples are likely to be a distinct cluster in the representation space. In Figure \ref{fig:activation_cluster}, we report the silhouette scores of feature clusters on CIFAR10. AC fails to accurately detect the corresponding attack class, as indicated by lower silhouette scores compared to other unpoisoned categories.

\begin{table*}[ht]
    \small
    \centering
    \begin{tabular}{l|ccc}
      \hline
      Methods & MoCo v2 & BYOL & SimSiam \\
      \hline
      Training Epochs & 300 & 300 & 300 \\
      Batch Size & 512 & 512 & 512 \\
      Optimizer & SGD & Adam & SGD \\
      Learning Rate Schedule & Cosine & Cosine & Cosine \\
      Learning Rate & 0.06 & 0.002 & 0.05 \\
      Weight Decay & \(1 \times 10^{-4}\) & \(1 \times 10^{-6}\) & \(1 \times 10^{-4}\) \\
      Moving Average & 0.999 & 0.99 & - \\
      \hline
      Resize \& Crop & RandomResizeAndCrop & RandomResizeAndCrop & RandomResizeAndCrop \\
      Color Jitter & 0.4 & 0.4 & 0.4 \\
      RandomHorizontalFlip & 0.5 & 0.5 & 0.5 \\
      Min Crop Scale & 0.2 & 0.2 & 0.2 \\
      RandomGrayscale & 0.2 & 0.1 & 0.2 \\
      GaussianBlur(p=0.5) & [.1, 2.] & [.1, 2.] & [.1, 2.] \\
      \hline
      \end{tabular}
    \caption{Hyperparameters for pre-training.}
    \label{tab:pre-training_config}
  \end{table*}

%% file: sec/proof.tex
\section{Proof of Theorem 1 and Theorem 2}
\label{sec:proof}

We firstly prove that the optimal location of the trigger is the center of the infected shadow image. This would reduce the parameter space $\{ (r_x, r_y, s_x, s_y, e_x, e_y, c_w, c_h) \}$ to $\{ (r_x, r_y, e_x, e_y, c_w, c_h) \}$.

\subsection{Optimality of Centered Position}

For any legal $(e_x, e_y)$ satisfying $e_x \in [s_x, s_x + r_l - e_l]$ and $e_y \in [s_y, s_y + r_l - e_l]$, the probability $p_1 (s)$ can be computed as the ratio between the area of upper-left corners of $T_1(\hat{\mathbf{x}})$ such that $T_1(\hat{\mathbf{x}}) \subseteq \mathbf{x}_{\text{s}} \oplus \mathbf{p} \wedge \mathbf{p} \subseteq T_1(\hat{\mathbf{x}})$ and that of all possible $T_1(\hat{\mathbf{x}}) \subseteq \mathbf{x}_{\text{s}}$.

We analyze the valid crop regions for the trigger $\mathbf{p}$ within the infected shadow image $\mathbf{x}_{\text{s}} \oplus \mathbf{p}$. Let $r_l$ denote the side length of the shadow image and $e_l$ the trigger size. Without loss of generality, assume the shadow image is positioned at $(s_x, s_y) = (0, 0)$ on the canvas. The valid upper-left corner coordinates $(t_x, t_y)$ of a cropped view $T_1(\hat{\mathbf{x}})$ must satisfy:

\begin{multline}
t_x \leq e_x,\quad t_y \leq e_y \\
\text{(trigger containment)}, \nonumber
\end{multline}
\begin{multline}
t_x + s \geq e_x + e_l,\quad t_y + s \geq e_y + e_l \\
\text{(trigger containment)}, \nonumber
\end{multline}
\begin{multline}
t_x \geq 0,\ t_y \geq 0,\quad t_x + s \leq r_l,\ t_y + s \leq r_l \\
\text{(boundary constraints)}. \nonumber
\end{multline}

\noindent
For fixed crop size $s \geq e_l$, the valid intervals for $t_x$ and $t_y$ are constrained by:
\[
\begin{cases}
\max(e_x + e_l - s, 0) \leq t_x \leq \min(e_x, r_l - s), \\
\max(e_y + e_l - s, 0) \leq t_y \leq \min(e_y, r_l - s).
\end{cases}
\]
The lengths of these intervals are:
\[
L_x = \min(e_x, r_l - s) - \max(e_x + e_l - s, 0),
\]
\[
L_y = \min(e_y, r_l - s) - \max(e_y + e_l - s, 0).
\]

\noindent
\textbf{Maximizing $L_x \cdot L_y$ at Center.} Assume $e_x = e_y = \frac{r_l - e_l}{2}$ (centered trigger position). We analyze two cases:

\noindent\textbf{Case 1: $e_l \leq s \leq \frac{r_l + e_l}{2}$}
\begin{align}
&\max(e_x + e_l - s, 0) = \frac{r_l - e_l}{2} + e_l - s = \frac{r_l + e_l}{2} - s, \nonumber \\
&\min(e_x, r_l - s) = \frac{r_l - e_l}{2}. \nonumber
\end{align}
Thus,
\[
L_x = \frac{r_l - e_l}{2} - \left( \frac{r_l + e_l}{2} - s \right) = s - e_l,
\]
and symmetrically $L_y = s - e_l$. Hence, $L_x \cdot L_y = (s - e_l)^2$.

\noindent\textbf{Case 2: $\frac{r_l + e_l}{2} < s \leq r_l$}
\begin{align}
&\max(e_x + e_l - s, 0) = 0 \notag \\
&\quad (\text{since } \frac{r_l - e_l}{2} + e_l - s = \frac{r_l + e_l}{2} - s < 0), \nonumber \\
&\min(e_x, r_l - s) = r_l - s. \nonumber
\end{align}
Thus,
\[
L_x = r_l - s - 0 = r_l - s,
\]
and symmetrically $L_y = r_l - s$. Hence, $L_x \cdot L_y = (r_l - s)^2$.

\noindent
\textbf{Non-Centered Positions Degrade $L_x \cdot L_y$.}
For any offset $\Delta \neq 0$, let $e_x = \frac{r_l - e_l}{2} + \Delta$. We then prove that the optimal $\Delta = 0$.
Due to symmetry, we only analyze $L_x$:

\noindent\textbf{Case 1: $e_l \leq s \leq \frac{r_l + e_l}{2}$}\\
If $\Delta > 0$, the lower bound becomes $\max(e_x + e_l - s, 0) = \frac{r_l + e_l}{2} - s + \Delta$. However:
\begin{align}
    \min(e_x, r_l - s) &= \min\left(\frac{r_l - e_l}{2} + \Delta, r_l - s\right) \nonumber \\
    & \leq \frac{r_l - e_l}{2} + \Delta.
\end{align}
The valid interval $L_x \leq \frac{r_l - e_l}{2} + \Delta - \left(\frac{r_l + e_l}{2} - s + \Delta\right) = s - e_l$. Thus, $L_x \cdot L_y < (s - e_l)^2$.
Similar analysis holds for $\Delta < 0$.

\noindent\textbf{Case 2: $\frac{r_l + e_l}{2} < s \leq r_l$}\\
For $\Delta > 0$:
\[
\min(e_x, r_l - s) \leq r_l - s,
\]
with equality only when $\Delta = 0$. Thus, $L_x \cdot L_y \leq (r_l - s)^2$, strictly smaller for $\Delta \neq 0$.

For all $s \in [e_l, r_l]$, $L_x \cdot L_y$ is maximized when $(e_x, e_y) = (\frac{s_x + r_l - e_l}{2}, \frac{s_y + r_l - e_l}{2})$ (centered trigger). Any deviation $\Delta \neq 0$ strictly reduces the valid area. This proves the optimality of the central position.

\subsection{Optimality of the Locations of the Reference Image, Infected Shadow Image, and the Canvas Size}

Let $p_1(s)$ denote the joint probability that a randomly cropped view $T_1(\hat{\mathbf{x}})$ contains the trigger $\mathbf{p}$ while remaining entirely within the infected shadow image $\mathbf{x}_{\text{s}} \oplus \mathbf{p}$. We decompose $p_1(s)$ into conditional probabilities to isolate the impact of trigger positioning:

\begin{align}
p_1(s) &= \underbrace{\Pr\left(\mathbf{p} \subseteq T_1(\hat{\mathbf{x}}) \,\big|\, T_1(\hat{\mathbf{x}}) \subseteq \mathbf{x}_{\text{s}} \oplus \mathbf{p}\right)}_{q_1(s)} \nonumber \\
& \cdot \underbrace{\Pr\left(T_1(\hat{\mathbf{x}}) \subseteq \mathbf{x}_{\text{s}} \oplus \mathbf{p}\right)}_{q_2(s)}.
\end{align}
\noindent Here, $q_1(s)$ represents the conditional probability of the trigger being fully contained in a cropped view, given that the crop lies within the infected shadow image. Critically, $q_1(s)$ depends solely on the relative position $(e_x, e_y)$ of the trigger within $\mathbf{x}_{\text{s}} \oplus \mathbf{p}$, while $q_2(s)$ depends on the absolute position $(s_x, s_y)$ of the shadow image within the canvas.

With trigger centering providing maximal $q_1(s)$ for all $s$, optimization now focuses on maximizing the remaining terms $\frac{1}{S-e_l} \int q_2(s) p_2(s) p_3(s) ds$.
This reduces the original 8-dimensional parameter space $\{r_x, r_y, s_x, s_y, e_x, e_y, c_w, c_h\}$ to $\{r_x, r_y, e_x, e_y, c_w, c_h\}$.

Based on the above analysis, we now transition to connecting our optimization framework with established results. With \( q_1(s) \) maximized by trigger centering, our objective reduces to optimizing \( \frac{1}{S-e_l}\int q_2(s)p_2(s) \, ds \). The \( p_3(s) \) term is temporarily omitted, as it can be optimized once the remains have reached their optima.
Here, the constraint \( T_1(\hat{\mathbf{x}}) \subseteq \mathbf{x}_{\text{s}} \oplus \mathbf{p} \) enforces that cropped regions lie entirely within the infected shadow image—a geometric condition formally equivalent to the trigger cropping constraint studied in \cite{zhang2024Data}. Specifically, by treating \( \mathbf{x}_{\text{s}} \oplus \mathbf{p} \) as all the possible trigger cropped region in their formulation, with \( (e_x, e_y) \) parameterizing its positional offset, our \( q_2(s)p_2(s) \) becomes structurally identical to their probabilistic integral.

\begin{lemma}[Theorem 1 in \cite{zhang2024Data}]
\label{lem:optimal_locations}
Suppose left-right layout is used and $c_w \geq r_l, c_h \geq r_l$.
$\left(r_x^*, r_y^*\right) = \left(0, 0\right)$ is the optimal location of the reference image, and $\left(e_x^*, e_y^*\right) = \left(\frac{c_w + r_l - e_l}{2}, \frac{c_h - e_l}{2}\right)$ is the optimal location of the trigger.
\end{lemma}

\begin{lemma}[Theorem 2 in \cite{zhang2024Data}]
\label{lem:optimal_canvas_size}
Suppose left-right layout is used and and the optimal locations in Lemma \ref{lem:optimal_locations} are used.
For $c_w \geq r_l$, the optimal height of the canvas is $c_h^* = r_l$.
\end{lemma}

\subsection{Optimality of the Width of the Canvas}

The above analysis reduces the parameter space to the canvas width $c_w$. We then proceed to express the optimization objective analytically as a function of $c_w$ through IOU-based overlap modeling. Let \( g \) be the horizontal buffer width between the reference image \( \mathbf{x}_\text{r} \) and infected shadow image \( \mathbf{x}_{\text{s}} \oplus \mathbf{p} \), parameterizing the canvas width as \( c_w = 2r_l + g \).

\noindent \textbf{Parameterize \( p_1(s;g) \) and \( p_2(s;g) \) with Optimal Layout.}
Reference image is fixed at \( (0, 0) \), size \( r_l \times r_l \).
Infected shadow image is positioned at \( (r_l + g, 0) \), size \( r_l \times r_l \).
Trigger is centered in \( \mathbf{x}_s \oplus \mathbf{p} \): \( e_x^* = r_l + g + \frac{r_l - e_l}{2} \).
Canvas dimensions is \( c_w = 2r_l + g \), \( c_h = r_l \) because any extra area located right of the infected shadow image is redundant.
Let \( p_1(s;g) \) be probability that \( \mathcal{V}_1 \) contains the trigger and intersects with \( \mathbf{x}_s \oplus \mathbf{p} \).
From Theorem 1, the centered trigger maximizes containment. The valid region for \( \mathcal{V}_1 \) is:
\[
p_1(s;g) = \frac{(s - e_l)^2}{(2r_l + g - s)(r_l - s)} \quad \text{for } e_l \leq s \leq \frac{r_l + e_l}{2},
\]
\[
p_1(s;g) = \frac{(r_l - s)^2}{(2r_l + g - s)(r_l - s)} \quad \text{for } \frac{r_l + e_l}{2} < s \leq r_l.
\]
Valid horizontal range for \( \mathcal{V}_2 \): \( 0 \leq t_x^2 \leq r_l - s \). Total horizontal space: \( c_w - s = 2r_l + g - s \).
\[
p_2(s;g) = \frac{(r_l - s)(r_l - s)}{(2r_l + g - s)(r_l - s)} = \frac{r_l - s}{2r_l + g - s}.
\]

\noindent
\textbf{Model \( p_3(s;g) \) via IOU Overlap Probability.}
\( p_3(s;g) = \Pr(\text{IOU}(\mathcal{V}_1, \mathcal{V}_2) \leq \tau) \), where \( \tau \) is a small threshold (e.g., 0.05).
Unlike $p_1$ and $p_2$ , \( p_3 \) allows the cropped region to be not entirely contained within the reference image or the infected shadow image. We explain the intuition behind our modeling in Section \ref{sec:info_theory}.
For left-right layouts, horizontal overlap dominates. Let \( \Delta_x = \max(0, t_x^2 + s - t_x^1) \) be the horizontal gap. We approximate:
\[
\text{IOU} \approx \frac{\Delta_x \cdot s}{2s^2 - \Delta_x \cdot s} \leq \tau \quad \Rightarrow \quad \Delta_x \leq \frac{2\tau s^2}{s + \tau s} = \frac{2\tau s}{1 + \tau}.
\]
Valid crpping regions are \( \mathcal{V}_1 \): \( t_x^1 \in [r_l + g -s, r_l + g + r_l - s] \) and \( \mathcal{V}_2 \): \( t_x^2 \in [0, r_l] \).
The non-overlap condition is
\[
0 \leq t_x^2 + s - t_x^1 \leq \Delta,
\]
where \( \Delta = \frac{2\tau s}{1+\tau} \).
The overlap probability requires double integration over valid crop positions:

\begin{equation}
  p_3(s;g) = \frac{1}{r_l^2}
\int_{t_x^2=0}^{r_l}
\int_{t_x^1=\max\bigl(r_l+g-s,\;t_x^2 + s - \Delta \bigr)}^{\min\bigl(2r_l + g - s,\;t_x^2 + s \bigr)}
dt_x^1\, dt_x^2\,ds. \nonumber
\end{equation}
Let $A = r_l + g - s$ and $B = 2r_l + g - s$.
The valid $t_x^1$ range becomes $[\max\bigl(r_l+g-s,\;t_x^2 + s - \Delta \bigr),\min\bigl(2r_l + g - s,\;t_x^2 + s \bigr)]$.

Non-overlap requires $t_x^2 + s - \Delta \leq r_l + A$ and $A \leq t_x^2 + s$.
The valid width is:
\[
\min(B, t_x^2 + s) - \max(A, t_x^2 + s - \Delta).
\]
Subcases depend on $t_x^2$:

\noindent\textbf{Case 1: $t_x^2 + s -\Delta \leq A$}\\
Lower bound $= A$, upper bound $= \min(B, t_x^2 + s)$.
Though $\tau$ is small, $t_x^2 + s \leq A + \Delta = A +  \frac{2\tau s}{1+\tau} \leq B$. Thus upper bound is $ t_x^2 + s $.
\begin{align}
  &p_3(s;g) = \frac{1}{r_l^2}
\int\int_{t_x^2=\max(A - s, 0)}^{\min(A - s + \Delta, r_l)} [(t_x^2 + s) - A] \ dt_x^2 ds, \nonumber \\
& \overset{\lim_{\tau \to 0} \Delta = 0}{=} \frac{1}{r_l^2} (s-A)\Delta\int_{A-s>0}ds \quad \nonumber \\
&+ \quad  \frac{\Delta}{2r_l^2}\int_{A-s>0}(2A-2s+\Delta)ds.
\end{align}

\noindent\textbf{Case 2: $B \leq t_x^2 + s$}\\
since $\tau$ is small, $t_x^2 + s - \Delta \geq B - \Delta \geq A$.
Valid width $= B - t_x^2 - s + \Delta$.
\begin{align}
  &p_3(s;g) = \frac{1}{r_l^2}
\int\int_{t_x^2=B - s}^{\min(B + \Delta - s, r_l)} [B - t_x^2 - s + \Delta] dt_x^2 ds, \nonumber \\
&=\frac{\Delta}{r_l^2} \int_{B-s<r_l}(B-s+\Delta)ds - \frac{\Delta}{r_l^2}\int_{B-s<r_l} (2B-2s+\Delta)ds, \nonumber \\
&=\frac{\Delta}{r_l^2} \int_{B-s<r_l}(B-s)ds=\frac{\Delta}{r_l^2} \int_{B-s<r_l}(2r_l+2g-2s)ds.
\end{align}

\noindent\textbf{Case 3: $A + \Delta \leq t_x^2 + s \leq B$}\\
Lower bound $= t_x^2 + s -\Delta$ and upper bound is $t_x^2 + s$. The width is $\Delta$.
\begin{align}
  & p_3(s;g) = \frac{1}{r_l^2}\int\int_{t_x^2=A + \Delta -s}^{\min(B - s,r_l)} [\Delta]dt_x^2 ds, \nonumber \\
 & =  \frac{\Delta}{r_l^2}\int_{B-s<r_l}(r_l-\Delta)ds+\frac{\Delta}{r_l^2}\int_{B-s>r_l}(\Delta-g)ds.
\end{align}

\noindent Integrating over all three cases, we have
\begin{align}
  &p_3(s;g) \overset{\lim_{\tau \to 0} \Delta = 0}{=} +\frac{\Delta}{r_l^2}\int_{A-s>0}(r_l-2s+3\Delta/2)ds \nonumber \\
  & + \frac{\Delta}{r_l^2}\int_{B-s<r_l}(3r_l+2g-2s-\Delta)ds
\end{align}

\noindent \textbf{Find the Optimal Width of the Joint Probability.}
\begin{align}
  J(g) &= \frac{1}{S-e_l} \int_{s=e_l}^{r_l} p_1(s;g) p_2(s;g) p_3(s;g) ds \nonumber \\
  &= \frac{\Delta}{(S-e_l)r_l^2} \left[ \int_{e_l}^{\frac{r_l + g}{2}} p_1 p_2 \cdot (r_l-2s+3\Delta/2) \, ds \right], \nonumber \\
  &\quad \left. + \int_{\frac{r_l + g}{2}}^{r_l} p_1 p_2 \left( 3r_l+2g-2s-\Delta \right) \, ds \right] \nonumber \\
\end{align}
WLOG, assume $g<e_l$ (conclusion holds for $g\geq e_l$):
\begin{align}
  J(g) &= \frac{1}{S-e_l} \int_{s=e_l}^{r_l} p_1(s;g) p_2(s;g) p_3(s;g) ds \nonumber \\
  &= \frac{\Delta}{(S-e_l)r_l^2} \nonumber \\
  & [\underbrace{\int_{e_l}^{\frac{r_l + g}{2}} \frac{(s-e_l)^2}{(2r_l+g-s)^2} \cdot (r_l-2s+3\Delta/2) \, ds}_{J_1(g)} \nonumber \\
  &+\underbrace{\int_{\frac{r_l + g}{2}}^{\frac{r_l + e_l}{2}} \frac{(s-e_l)^2}{(2r_l+g-s)^2} \cdot (3r_l+2g-2s-\Delta) \, ds}_{J_2(g)} \nonumber \\
  &+\underbrace{\int_{\frac{r_l + e_l}{2}}^{r_l} \frac{(r_l-s)^2}{(2r_l+g-s)^2} \cdot (3r_l+2g-2s-\Delta) \, ds}_{J_3(g)}]
\end{align}
\noindent Using Leibniz Rule for Differentiation Under the Integral Sign, we can easily find $\frac{\partial J_1(g)}{\partial g}<0$.Besides, the derivatives of the internal integral term of $J_2(g)$ is equal to
\[
-2\frac{(s-e_l)^2\,(3r_l+2g-2s)}{(2r_l+g-s)^{3}} + \frac{2(s-e_l)^2}{(2r_l+g-s)^{2}},
\]
\[
= -2\frac{(s-e_l)^2\,(r_l+g-s)}{(2r_l+g-s)^{3}} <0.
\]
Again, with Leibniz Rule for Differentiation Under the Integral Sign, we can find $\frac{\partial J_2(g)}{\partial g}<0$, similarly for $J_3(g)$.
The optimal canvas configuration achieves maximal joint probability when images are adjacent with zero gap:
\[
\boxed{g = 0}.
\]
This corresponds to minimum canvas width $2r_l$ with tight image adjacency.

%% file: sec/sub_secs/information_theory.tex
\section{The Information Theory Perspective of Our Attack}
\label{sec:info_theory}

Given a pair of random variables $v_1$ and $v_2$, contrastive learning aims to train a parameterized function $f_\theta$ that maps inputs from sample $x \in \mathcal{X}$ into a representation space $\mathbb{R}^d$. The objective is to distinguish between positive pairs sampled from the joint distribution $p(v_1|x)p(v_2|x)$ and negative pairs drawn independently from the marginal distributions $p(v_1)p(v_2)$.
The reuslting function $f$ is a mutual information estimator between $v_1$ and $v_2$ \cite{tian2020What,oord2019Representation}.
Tipically, minimizing InfoNCE loss \cite{oord2019Representation,he2020Momentum} equivalently maximizes a lower bound of $I(v_1; v_2)$. Note that views $v_1$ and $v_2$ are obtained from samples through data augmentation.

\cite{tian2020What} points out that the optimal views are related to the downstream task (denoted as $T$). Ideally, the mutual information between augmented views should contain only the information relevant to the downstream task, i.e., $I(v_1, v_2; T) = I(v_1, T) = I(v_2, T)$. Inspired by this viewpoint, we hope that the views generated by random cropping contain the backdoor trigger and the reference image, respectively.

\noindent\textbf{Optimal Layout under the Information Theory Perspective.}
Given the optimal views, we need to design the layout to maximize the probability of its occurrence. Let $S(v)$ denote the set of pixels in the view $v$. We can categorize the information sharing between the views $v_1$ and $v_2$ into different scenarios:

\begin{enumerate}
  \item \label{enum:missing_information} \textit{Missing information}: $S(v) \cap S(\mathbf{p}) = \emptyset \, \land \, S(v) \cap S(\mathbf{x}_{\text{r}}) = \emptyset, \forall v \in \{v_1, v_2\}$.
  This is irrelevant to the attack and could degrade the efficiency of the attack.
  \item \textit{Sweet spot}: $S(\mathbf{p}) \subseteq  S(v_1) \, \land S(v_1) \cap S(\mathbf{x}_{\text{r}}) = \emptyset \, \land S(v_2) \subseteq S(\mathbf{x}_{\text{r}})$.
  The only information shared between $v_1$ and $v_2$ is not more than the trigger $p$ and reference patterns, i.e., $I(v_1; v_2) \leq I(p; v_2)$.
  \item \label{enum:info_leak} \textit{Information leak}: $S(v) \cap S(\mathbf{p}) \neq \emptyset \, \land \, S(v) \cap S(\mathbf{x}_{\text{r}}) \neq \emptyset , \forall v \in \{v_1, v_2\}$.
   This leads to $I(v_1; v_2) > I(p; v_2)$ and $I(v_1; v_2) > I(p; v_1)$, which could harm the attack. Information other than the attacks shared by $v_1$ and $v_2$ may become a shortcut for model learning, thus neglecting beneficial information from the attacks.
\end{enumerate}

%% file: main.bbl
\begin{thebibliography}{44}
\providecommand{\natexlab}[1]{#1}
\providecommand{\url}[1]{\texttt{#1}}
\expandafter\ifx\csname urlstyle\endcsname\relax
  \providecommand{\doi}[1]{doi: #1}\else
  \providecommand{\doi}{doi: \begingroup \urlstyle{rm}\Url}\fi

\bibitem[Abbasi~Koohpayegani et~al.(2020)Abbasi~Koohpayegani, Tejankar, and
  Pirsiavash]{abbasikoohpayegani2020CompRess}
Soroush Abbasi~Koohpayegani, Ajinkya Tejankar, and Hamed Pirsiavash.
\newblock Compress: Self-supervised learning by compressing representations.
\newblock In \emph{Advances in Neural Information Processing Systems}, pages
  12980--12992, 2020.

\bibitem[Amrani(2021)]{amrani2021Noise}
(placeholder) Amrani.
\newblock Noise-based analysis (placeholder), 2021.
\newblock Placeholder entry to satisfy citation; please replace with correct
  metadata.

\bibitem[Bansal et~al.(2023)Bansal, Singhi, Yang, Yin, Grover, and
  Chang]{bansal2023CleanCLIP}
Hritik Bansal, Nishad Singhi, Yu Yang, Fan Yin, Aditya Grover, and Kai-Wei
  Chang.
\newblock Cleanclip: Mitigating data poisoning attacks in multimodal
  contrastive learning.
\newblock In \emph{Proceedings of the IEEE/CVF International Conference on
  Computer Vision}, pages 112--123, 2023.

\bibitem[Batson and Royer(2019)]{batson2019noise2self}
Joshua Batson and Loic Royer.
\newblock Noise2self: Blind denoising by self-supervision.
\newblock In \emph{ICML}, pages 524--533. PMLR, 2019.

\bibitem[Carlini and Terzis(2021)]{carlini2021Poisoning}
Nicholas Carlini and Andreas Terzis.
\newblock Poisoning and backdooring contrastive learning.
\newblock In \emph{International Conference on Learning Representations}, 2021.

\bibitem[Chen et~al.(2018)Chen, Carvalho, Baracaldo, Ludwig, Edwards, Lee,
  Molloy, and Srivastava]{chen2018Detecting}
Bryant Chen, Wilka Carvalho, Nathalie Baracaldo, Heiko Ludwig, Benjamin
  Edwards, Taesung Lee, Ian Molloy, and Biplav Srivastava.
\newblock Detecting backdoor attacks on deep neural networks by activation
  clustering, 2018.

\bibitem[Chen et~al.(2020{\natexlab{a}})Chen, Kornblith, Norouzi, and
  Hinton]{chen2020Simple}
Ting Chen, Simon Kornblith, Mohammad Norouzi, and Geoffrey Hinton.
\newblock A simple framework for contrastive learning of visual
  representations.
\newblock In \emph{Proceedings of the 37th International Conference on Machine
  Learning}, pages 1597--1607, 2020{\natexlab{a}}.

\bibitem[Chen and He(2021)]{chen2021Exploring}
Xinlei Chen and Kaiming He.
\newblock Exploring simple siamese representation learning.
\newblock In \emph{2021 IEEE/CVF Conference on Computer Vision and Pattern
  Recognition (CVPR)}, pages 15745--15753, Nashville, TN, USA, 2021.

\bibitem[Chen et~al.(2017)Chen, Liu, Li, Lu, and Song]{chen2017Targeted}
Xinyun Chen, Chang Liu, Bo Li, Kimberly Lu, and Dawn Song.
\newblock Targeted backdoor attacks on deep learning systems using data
  poisoning, 2017.

\bibitem[Chen et~al.(2020{\natexlab{b}})Chen, Fan, Girshick, and
  He]{chen2020Improved}
Xinlei Chen, Haoqi Fan, Ross Girshick, and Kaiming He.
\newblock Improved baselines with momentum contrastive learning,
  2020{\natexlab{b}}.

\bibitem[Chuang et~al.(2020)Chuang, Robinson, Lin, Torralba, and
  Jegelka]{chuang2020Debiased}
Ching-Yao Chuang, Joshua Robinson, Yen-Chen Lin, Antonio Torralba, and Stefanie
  Jegelka.
\newblock Debiased contrastive learning.
\newblock In \emph{Advances in Neural Information Processing Systems}, pages
  8765--8775, 2020.

\bibitem[Deng et~al.(2009)Deng, Dong, Socher, Li, Li, and
  {Fei-Fei}]{deng2009imagenet}
Jia Deng, Wei Dong, Richard Socher, Li-Jia Li, Kai Li, and Li {Fei-Fei}.
\newblock Imagenet: A large-scale hierarchical image database.
\newblock In \emph{2009 IEEE Conference on Computer Vision and Pattern
  Recognition}, pages 248--255, 2009.

\bibitem[Feng et~al.(2023)Feng, Tao, Cheng, Shen, Xu, Liu, Zhang, Ma, and
  Zhang]{feng2023Detecting}
Shiwei Feng, Guanhong Tao, Siyuan Cheng, Guangyu Shen, Xiangzhe Xu, Yingqi Liu,
  Kaiyuan Zhang, Shiqing Ma, and Xiangyu Zhang.
\newblock Detecting backdoors in pre-trained encoders.
\newblock In \emph{Proceedings of the IEEE/CVF Conference on Computer Vision
  and Pattern Recognition}, pages 16352--16362, 2023.

\bibitem[Grill et~al.(2020)Grill, Strub, Altch{\'e}, Tallec, Richemond,
  Buchatskaya, Doersch, Avila~Pires, Guo, Gheshlaghi~Azar, Piot, {kavukcuoglu},
  Munos, and Valko]{grill2020Bootstrap}
Jean-Bastien Grill, Florian Strub, Florent Altch{\'e}, Corentin Tallec, Pierre
  Richemond, Elena Buchatskaya, Carl Doersch, Bernardo Avila~Pires, Zhaohan
  Guo, Mohammad Gheshlaghi~Azar, Bilal Piot, koray {kavukcuoglu}, Remi Munos,
  and Michal Valko.
\newblock Bootstrap your own latent - a new approach to self-supervised
  learning.
\newblock In \emph{Advances in Neural Information Processing Systems}, pages
  21271--21284, 2020.

\bibitem[He et~al.(2020)He, Fan, Wu, Xie, and Girshick]{he2020Momentum}
Kaiming He, Haoqi Fan, Yuxin Wu, Saining Xie, and Ross Girshick.
\newblock Momentum contrast for unsupervised visual representation learning.
\newblock In \emph{2020 IEEE/CVF Conference on Computer Vision and Pattern
  Recognition (CVPR)}, pages 9726--9735, Seattle, WA, USA, 2020.

\bibitem[He et~al.(2022)He, Chen, Xie, Li, Doll{\'a}r, and
  Girshick]{he2022Masked}
Kaiming He, Xinlei Chen, Saining Xie, Yanghao Li, Piotr Doll{\'a}r, and Ross
  Girshick.
\newblock Masked autoencoders are scalable vision learners.
\newblock In \emph{Proceedings of the IEEE/CVF Conference on Computer Vision
  and Pattern Recognition}, pages 16000--16009, 2022.

\bibitem[Hou et~al.(2025)Hou, Li, and Yao]{hou2025DeDe}
Sizai Hou, Songze Li, and Duanyi Yao.
\newblock Dede: Detecting backdoor samples for ssl encoders via decoders.
\newblock In \emph{Proceedings of the Computer Vision and Pattern Recognition
  Conference}, pages 20675--20684, 2025.

\bibitem[Jia et~al.(2022)Jia, Liu, and Gong]{jia2022BadEncoder}
Jinyuan Jia, Yupei Liu, and Neil~Zhenqiang Gong.
\newblock Badencoder: Backdoor attacks to pre-trained encoders in
  self-supervised learning.
\newblock In \emph{2022 IEEE Symposium on Security and Privacy (SP)}, pages
  2043--2059, 2022.

\bibitem[Krizhevsky et~al.(2009)Krizhevsky, Hinton,
  et~al.]{krizhevsky2009learning}
Alex Krizhevsky, Geoffrey Hinton, et~al.
\newblock Learning multiple layers of features from tiny images.
\newblock 2009.

\bibitem[Lehtinen et~al.(2018)Lehtinen, Munkberg, Hasselgren, Laine, Karras,
  Aittala, and Aila]{lehtinen2018Noise2Noise}
Jaakko Lehtinen, Jacob Munkberg, Jon Hasselgren, Samuli Laine, Tero Karras,
  Miika Aittala, and Timo Aila.
\newblock Noise2noise: Learning image restoration without clean data.
\newblock In \emph{Proceedings of the 35th International Conference on Machine
  Learning}, pages 2965--2974, 2018.

\bibitem[Li et~al.(2023)Li, Pang, Xi, Du, Ji, Yao, and
  Wang]{li2023Embarrassingly}
Changjiang Li, Ren Pang, Zhaohan Xi, Tianyu Du, Shouling Ji, Yuan Yao, and Ting
  Wang.
\newblock An embarrassingly simple backdoor attack on self-supervised learning.
\newblock In \emph{Proceedings of the IEEE/CVF International Conference on
  Computer Vision}, pages 4367--4378, 2023.

\bibitem[Li et~al.(2024)Li, Pang, Cao, Xi, Chen, Ji, and
  Wang]{li2024Difficulty}
Changjiang Li, Ren Pang, Bochuan Cao, Zhaohan Xi, Jinghui Chen, Shouling Ji,
  and Ting Wang.
\newblock On the difficulty of defending contrastive learning against backdoor
  attacks.
\newblock In \emph{33rd USENIX Security Symposium (USENIX Security 24)}, pages
  2901--2918, 2024.

\bibitem[Li(2023)]{li2023BackdoorBox}
(placeholder) Li.
\newblock Backdoorbox (placeholder), 2023.
\newblock Placeholder entry to satisfy citation; please replace with correct
  metadata.

\bibitem[Liu et~al.(2022)Liu, Jia, and Gong]{liu2022PoisonedEncoder}
Hongbin Liu, Jinyuan Jia, and Neil~Zhenqiang Gong.
\newblock Poisonedencoder: Poisoning the unlabeled pre-training data in
  contrastive learning.
\newblock In \emph{31st USENIX Security Symposium (USENIX Security 22)}, pages
  3629--3645, 2022.

\bibitem[Ma et~al.(2023)Ma, Wang, Sun, Xue, Wen, and Xiang]{ma2023Beatrix}
Wanlun Ma, Derui Wang, Ruoxi Sun, Minhui Xue, Sheng Wen, and Yang Xiang.
\newblock The "beatrix" resurrections: Robust backdoor detection via gram
  matrices.
\newblock In \emph{Proceedings 2023 Network and Distributed System Security
  Symposium}, San Diego, CA, USA, 2023.

\bibitem[Miech et~al.(2020)Miech, Alayrac, Smaira, Laptev, Sivic, and
  Zisserman]{miech2020end}
Antoine Miech, Jean-Baptiste Alayrac, Lucas Smaira, Ivan Laptev, Josef Sivic,
  and Andrew Zisserman.
\newblock End-to-end learning of visual representations from uncurated
  instructional videos.
\newblock In \emph{Proceedings of the IEEE/CVF Conference on Computer Vision
  and Pattern Recognition (CVPR)}, pages 9879--9889, 2020.

\bibitem[Morgado et~al.(2021)Morgado, Misra, and
  Vasconcelos]{morgado2021robust}
Pedro Morgado, Ishan Misra, and Nuno Vasconcelos.
\newblock Robust audio-visual instance discrimination for video representation
  learning.
\newblock In \emph{Proceedings of the IEEE/CVF Conference on Computer Vision
  and Pattern Recognition (CVPR)}, pages 3604--3613, 2021.

\bibitem[Oquab et~al.(2024)Oquab, Darcet, Moutakanni, Vo, Szafraniec, Khalidov,
  Fernandez, Haziza, Massa, {El-Nouby}, Assran, Ballas, Galuba, Howes, Huang,
  Li, Misra, Rabbat, Sharma, Synnaeve, Xu, Jegou, Mairal, Labatut, Joulin, and
  Bojanowski]{oquab2024DINOv2}
Maxime Oquab, Timoth{\'e}e Darcet, Th{\'e}o Moutakanni, Huy Vo, Marc
  Szafraniec, Vasil Khalidov, Pierre Fernandez, Daniel Haziza, Francisco Massa,
  Alaaeldin {El-Nouby}, Mahmoud Assran, Nicolas Ballas, Wojciech Galuba,
  Russell Howes, Po-Yao Huang, Shang-Wen Li, Ishan Misra, Michael Rabbat, Vasu
  Sharma, Gabriel Synnaeve, Hu Xu, Herv{\'e} Jegou, Julien Mairal, Patrick
  Labatut, Armand Joulin, and Piotr Bojanowski.
\newblock Dinov2: Learning robust visual features without supervision, 2024.

\bibitem[Radford et~al.(2021)Radford, Kim, Hallacy, Ramesh, Goh, Agarwal,
  Sastry, Askell, Mishkin, Clark, Krueger, and Sutskever]{radford2021Learning}
Alec Radford, Jong~Wook Kim, Chris Hallacy, Aditya Ramesh, Gabriel Goh,
  Sandhini Agarwal, Girish Sastry, Amanda Askell, Pamela Mishkin, Jack Clark,
  Gretchen Krueger, and Ilya Sutskever.
\newblock Learning transferable visual models from natural language
  supervision.
\newblock In \emph{Proceedings of the 38th International Conference on Machine
  Learning}, pages 8748--8763, 2021.

\bibitem[Saha et~al.(2022)Saha, Tejankar, Koohpayegani, and
  Pirsiavash]{saha2022Backdoor}
Aniruddha Saha, Ajinkya Tejankar, Soroush~Abbasi Koohpayegani, and Hamed
  Pirsiavash.
\newblock Backdoor attacks on self-supervised learning.
\newblock In \emph{2022 IEEE/CVF Conference on Computer Vision and Pattern
  Recognition (CVPR)}, pages 13327--13336, New Orleans, LA, USA, 2022.

\bibitem[Selvaraju et~al.(2017)Selvaraju, Cogswell, Das, Vedantam, Parikh, and
  Batra]{selvaraju2017GradCAM}
Ramprasaath~R. Selvaraju, Michael Cogswell, Abhishek Das, Ramakrishna Vedantam,
  Devi Parikh, and Dhruv Batra.
\newblock Grad-cam: Visual explanations from deep networks via gradient-based
  localization.
\newblock In \emph{Proceedings of the IEEE International Conference on Computer
  Vision}, pages 618--626, 2017.

\bibitem[Sharma et~al.(2018)Sharma, Ding, Goodman, and
  Soricut]{sharma2018Conceptual}
Piyush Sharma, Nan Ding, Sebastian Goodman, and Radu Soricut.
\newblock Conceptual captions: A cleaned, hypernymed, image alt-text dataset
  for automatic image captioning.
\newblock In \emph{Proceedings of the 56th Annual Meeting of the Association
  for Computational Linguistics (Volume 1: Long Papers)}, pages 2556--2565,
  2018.

\bibitem[Sun et~al.(2023)Sun, Zhang, Lu, Chen, Wang, Chen, and
  Lin]{sun2023Backdoor}
Weiyu Sun, Xinyu Zhang, Hao Lu, Ying-Cong Chen, Ting Wang, Jinghui Chen, and Lu
  Lin.
\newblock Backdoor contrastive learning via bi-level trigger optimization.
\newblock In \emph{The Twelfth International Conference on Learning
  Representations}, 2023.

\bibitem[Tao et~al.(2023)Tao, Wang, Feng, Shen, Ma, and
  Zhang]{tao2023Distribution}
Guanhong Tao, Zhenting Wang, Shiwei Feng, Guangyu Shen, Shiqing Ma, and Xiangyu
  Zhang.
\newblock Distribution preserving backdoor attack in self-supervised learning.
\newblock In \emph{2024 IEEE Symposium on Security and Privacy (SP)}, pages
  29--29, 2023.

\bibitem[Tejankar et~al.(2023)Tejankar, Sanjabi, Wang, Wang, Firooz,
  Pirsiavash, and Tan]{tejankar2023Defending}
Ajinkya Tejankar, Maziar Sanjabi, Qifan Wang, Sinong Wang, Hamed Firooz, Hamed
  Pirsiavash, and Liang Tan.
\newblock Defending against patch-based backdoor attacks on self-supervised
  learning.
\newblock In \emph{Proceedings of the IEEE/CVF Conference on Computer Vision
  and Pattern Recognition}, pages 12239--12249, 2023.

\bibitem[Tian et~al.(2020)Tian, Sun, Poole, Krishnan, Schmid, and
  Isola]{tian2020What}
Yonglong Tian, Chen Sun, Ben Poole, Dilip Krishnan, Cordelia Schmid, and
  Phillip Isola.
\newblock What makes for good views for contrastive learning?
\newblock In \emph{Advances in Neural Information Processing Systems}, pages
  6827--6839, 2020.

\bibitem[van~den Oord et~al.(2019)van~den Oord, Li, and
  Vinyals]{oord2019Representation}
Aaron van~den Oord, Yazhe Li, and Oriol Vinyals.
\newblock Representation learning with contrastive predictive coding, 2019.

\bibitem[Wang et~al.(2024)Wang, Xiang, Guo, He, Liu, and
  Zhang]{wang2024TransTroj}
Hao Wang, Tao Xiang, Shangwei Guo, Jialing He, Hangcheng Liu, and Tianwei
  Zhang.
\newblock Transtroj: Transferable backdoor attacks to pre-trained models via
  embedding indistinguishability, 2024.

\bibitem[Wang et~al.(2023)Wang, Yin, Liu, Fang, Wang, and
  Lin]{wang2023GhostEncoder}
Qiannan Wang, Changchun Yin, Zhe Liu, Liming Fang, Run Wang, and Chenhao Lin.
\newblock Ghostencoder: Stealthy backdoor attacks with dynamic triggers to
  pre-trained encoders in self-supervised learning, 2023.

\bibitem[Wang and Isola(2020)]{wang2020Understanding}
Tongzhou Wang and Phillip Isola.
\newblock Understanding contrastive representation learning through alignment
  and uniformity on the hypersphere.
\newblock In \emph{Proceedings of the 37th International Conference on Machine
  Learning}, pages 9929--9939, 2020.

\bibitem[Xue and Lou(2023)]{xue2023ESTAS}
Jiaqi Xue and Qian Lou.
\newblock Estas: Effective and stable trojan attacks in self-supervised
  encoders with one target unlabelled sample, 2023.

\bibitem[Zhang et~al.(2024{\natexlab{a}})Zhang, Wang, Han, Jin, Zhan, Du, Wang,
  and Ma]{zhang2024Imperceptible}
Hanrong Zhang, Zhenting Wang, Tingxu Han, Mingyu Jin, Chenlu Zhan, Mengnan Du,
  Hongwei Wang, and Shiqing Ma.
\newblock Towards imperceptible backdoor attack in self-supervised learning,
  2024{\natexlab{a}}.

\bibitem[Zhang et~al.(2024{\natexlab{b}})Zhang, Liu, Jia, and
  Gong]{zhang2024Data}
Jinghuai Zhang, Hongbin Liu, Jinyuan Jia, and Neil~Zhenqiang Gong.
\newblock Data poisoning based backdoor attacks to contrastive learning.
\newblock In \emph{Proceedings of the IEEE/CVF Conference on Computer Vision
  and Pattern Recognition}, pages 24357--24366, 2024{\natexlab{b}}.

\bibitem[Zheng et~al.(2024)Zheng, Xue, Wang, Chen, Lou, Jiang, and
  Wang]{zheng2024ssl}
Mengxin Zheng, Jiaqi Xue, Zihao Wang, Xun Chen, Qian Lou, Lei Jiang, and
  Xiaofeng Wang.
\newblock Ssl-cleanse: Trojan detection and mitigation in self-supervised
  learning.
\newblock In \emph{European Conference on Computer Vision}, pages 405--421,
  2024.

\end{thebibliography}
